\documentclass[10pt,twoside]{styles/csthesispaper}
\usepackage{styles/s5paper}
\usepackage[utf8]{inputenc}

\usepackage{natbib}
\PassOptionsToPackage{inline}{enumitem}

\usepackage{multirow}
\usepackage{tabularx}
\usepackage{xspace}
\usepackage{booktabs}
\usepackage{graphicx}
\usepackage[hidelinks]{hyperref}
\usepackage[noend]{algpseudocode}
\usepackage[linesnumbered,lined,ruled]{algorithm2e}
\usepackage{tikz}
\usepackage{subfig}
\usepackage{enumitem}
\usepackage{amsmath}
\usepackage{float} 
\usepackage{rotating}

\setlist{partopsep=0pt,parsep=0pt,topsep=3pt}
\setlist[description]{leftmargin=0pt,style=unboxed}

\newcommand*\mycircled[1]{\tikz[baseline=(char.base)]{\node[shape=circle,draw,inner sep=1pt] (char) {#1};}}

\newcommand{\FR}[2][]{{\advance\marginparsep by 2em\advance\marginparwidth by 2em\textcolor{red}{#1}\todo[color=yellow]{\scriptsize F: #2\\}}}

\newcommand{\ourmodel}{Cformer\xspace}
\newcommand{\ourmodelM}{CformerM\xspace}

\newcommand{\ag}{AG News\xspace}
\newcommand{\yahoo}{Yahoo!\ Answers\xspace}
\newcommand{\bonnier}{Bonnier News\xspace}

\newcolumntype{Y}{>{\centering\arraybackslash}X}

\newcommand{\norm}[1]{\|#1\|}

\DeclareMathOperator*{\argmax}{argmax}

\newcommand{\CUMass}{\ensuremath{C_{\textrm{UMass}}}\xspace}
\newcommand{\Cv}{\ensuremath{C_\textrm v}\xspace}

\newcommand{\data}[1]{\ensuremath{D_{\mathrm #1}}\xspace}
\newcommand{\Lab}{\data g}
\newcommand{\Ulab}{\data u}
\newcommand{\Plab}{\data p}
\newcommand{\Aug}{\data a}
\newcommand{\Wtar}{\ensuremath{W_\mathrm{target}}\xspace}
\newcommand{\Wsupp}{\ensuremath{W_\mathrm{supp}}\xspace}

\newcommand{\lab}{x_{\mathrm g}}

\newcommand{\plab}{x_{\mathrm p}}
\newcommand{\aug}{x_{\mathrm a}}
\newcommand{\lbl}[2][\relax]{%
    \def\tmp{#1}%
    \ifx\tmp\relax\ell(#2)%
    \else\ell_{#1}(#2)\fi%
}

\newcommand{\shrp}{\mathrm{sharp}}
\newcommand{\hrd}{\mathrm{hard}}
\newcommand{\shrpn}{\mathrm{sharpen}}
\newcommand{\Loss}{\mathrm{Loss}}
\newcommand{\mpl}{\mathrm{MPL}}

\newcommand{\question}[1]{Q$_{#1}$}

\title{The Efficiency of Pre-training with Objective Masking in Pseudo Labeling for Semi-Supervised Text Classification}

\begin{document}
\pagestyle{empty}

\author{Arezoo Hatefi, Xuan-Son Vu, Monowar Bhuyan, Frank Drewes}
\address{Department of Computing Science, Umeå University, Umeå, Sweden}
\address{arezooh@cs.umu.se, sonvx@cs.umu.se, monowar@cs.umu.se, drewes@cs.umu.se}

\maketitle
\thispagestyle{empty}

\abstract
We extend and study a semi-supervised model for text classification proposed earlier by Hatefi et al.\ for classification tasks in which document classes are described by a small number of gold-labeled examples, while the majority of training examples is unlabeled. The model leverages the teacher-student architecture of Meta Pseudo Labels in which a ``teacher'' generates labels for originally unlabeled training data to train the ``student'' and updates its own model iteratively based on the performance of the student on the gold-labeled portion of the data. 
We extend the original model of Hatefi et al.\ by an unsupervised pre-training phase based on objective masking, and conduct in-depth performance evaluations of the original model, our extension, and various independent baselines. Experiments are performed using three different datasets in two different languages (English and Swedish).

\section{Introduction}

Automatic topic classification of news articles is of great practical and commercial interest because of the huge number of news articles produced around the globe every day. \citet{Hatefi.etAl:21} have proposed a semi-supervised model \ourmodel for this task. One application area described at some length in that article is contextual advertising, also called cookieless advertising, which places ads in online news media based on the content of the news being viewed rather than on personal information about the viewer.

Applications such as this one often need to pre-determine the topics of interest. For example, a company running an advertisement campaign usually wants its ads to be seen in certain contexts and -- often more importantly -- not to be seen in others. For this, one would like to tag each article by a topic of interest. To create such a topic model, one would ideally want to train it on labeled data. However, since the topics of interest may frequently change and cannot be expected to be taken from a preconceived global set, the assumption of having sufficiently much labeled data for training is unrealistic. The best one may hope for is to be given a training set that consists of
\begin{enumerate*}[label=(\alph*)]
\item a comparatively large set \Ulab of unlabeled articles of the general type to be classified and
\item a much smaller set \Lab of gold-labeled examples for each of the topics in question.
\end{enumerate*}

Given the effectiveness of neural methods in natural language processing, a widespread approach to grouping documents into topics is to define a similarity measure via the distance between document embedding vectors and then use a general purpose clustering method such as $K$-Means to group documents into distinct classes \citep{zhang-etal-2022-neural,Grootendorst2022BERTopicNT}. However, as pointed out by~\citet{aggarwal2001surprising} the high dimensionality of these vector spaces has a negative impact on the accuracy of such an approach. Therefore, more robust methods such as MixText \citep{chen2020mixtext} and UDA \citep{xie2019unsupervised} have been proposed in the literature. A third approach, recently proposed by~\citet{Hatefi.etAl:21}, is called \ourmodel. It is a semi-supervised approach that makes use of a small set \Lab of gold-labeled documents (such as news articles provided as typical examples of the clusters of interest) and a larger set \Ulab of unlabeled documents. \ourmodel employs a student-teacher architecture originally proposed by \citet{pham2020meta} for computer vision: two BERT models, the \emph{teacher} and the \emph{student}, are trained in an iterated fashion. The teacher predicts pseudo-labels for documents in \Ulab, thus turning it into a pseudo-labeled dataset \Plab. The dataset $\Plab$ is then used for supervised training of the student. In the next step, the teacher's ability to predict pseudo-labels is improved, using the performance of the student on~\Lab as its objective function. Eventually, when the iterative process has converged, \Lab is used to fine-tune the student, yielding the final model. An advantage of \ourmodel is that the student can be replaced by one based on a smaller model such as DistilBERT to reduce the model size. This variant of \ourmodel is called Distil-\ourmodel. The empirical results of \citet{Hatefi.etAl:21} indicated that both \ourmodel and Distil-\ourmodel yield results on par with or better than MixText and UDA.

\begin{figure}[h!]
	\centering
	\includegraphics[width=\linewidth]{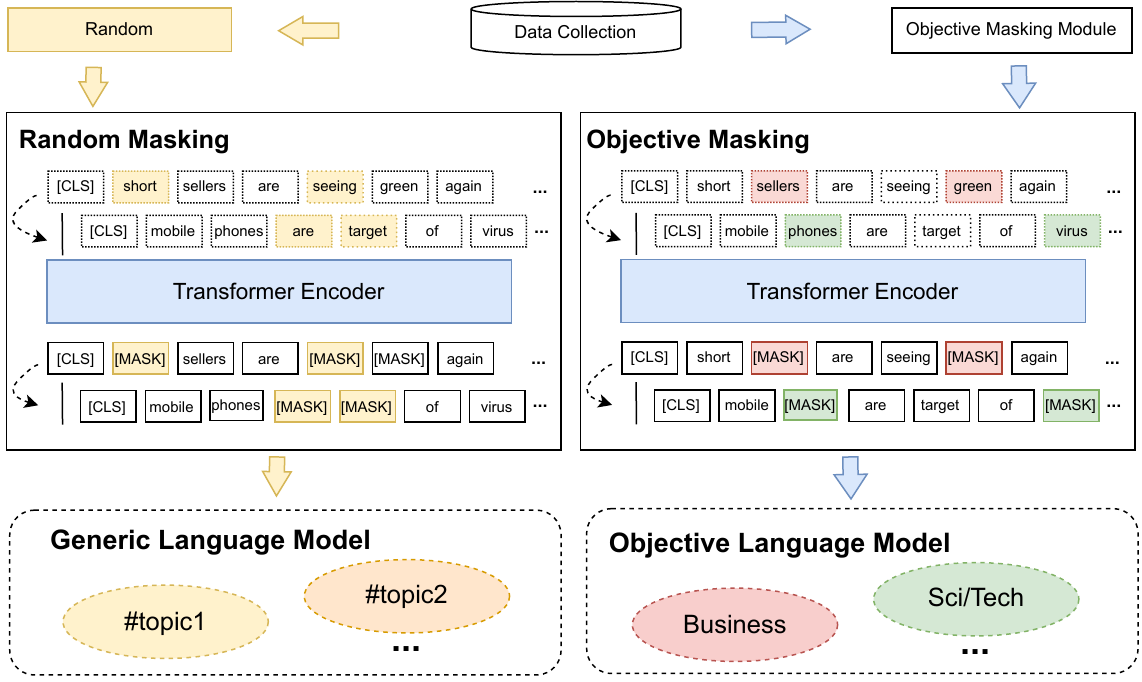}
	\caption{A high-level overview of \emph{Random Masking} in comparison to \emph{Objective Masking}. While the former is used for general purpose language models, the latter is preferable for increasing the sensitivity of a language model to topical information. 
	}
	\label{fig:architecture}
\end{figure}


The major contributions of this paper are
\begin{itemize}
    \item the model \ourmodelM that extends \ourmodel by including an unsupervised pre-training phase based on objective masking,
    \item a comparison of \ourmodel and \ourmodelM with each other as well as with MixText \citep{chen2020mixtext}, UDA \citep{xie2019unsupervised}, FLiText \citep{liu-etal-2021-flitext}, PGPL \citep{yang-etal-2023-prototype}, and BERT \citep{devlin2018bert} on \ag and \yahoo, Medical Abstracts, and a Swedish real-world dataset of news articles, and
    \item an investigation of the effectiveness of the objective masking in \ourmodelM, its impact on zero-shot classification, and its effect on the reliability and interpretability of the model.
\end{itemize}

The purpose of introducing masking into the learning process is to fine-tune the basic BERT models underlying the teacher and the student to improve their ability to recognize topical information in~$D$ (see Figure \ref{fig:architecture}). For this, we first create an independent unsupervised topic model for the dataset using Latent Dirichlet Allocation (LDA, \citet{944937}). This enables us to choose words carrying topical information. These words are used in objective masking to pre-train the BERT models which will afterwards be trained in the semi-supervised fashion of \ourmodel.

As we shall see, the model that uses objective masking, called \ourmodelM, outperforms \ourmodel and other SoTA baselines over two public benchmark datasets (in English) and one private dataset (in Swedish).\footnote{The private dataset is available from the owner upon request; see Section \ref{sec:conclusion}. This ensures that our results can be reproduced.} This confirms the effectiveness of objective masking in adjusting the language models (LMs) to the classification task. However, the extent of the improvement depends on the characteristics of the dataset and the number of labeled examples.


\noindent \textit{Organization.} The rest of the paper is organized as follows. An introduction to the basic notions used in this research and a discussion of related work is provided in Section \ref{sec:prelim}. A description of the architecture of \ourmodel and how it is turned into \ourmodelM can be found in Section \ref{sec:method}. The experimental setup and a comprehensive analysis of results are presented in Section \ref{sec:exp_analyse}. Finally, Section \ref{sec:conclusion} summarizes our main conclusions.

\section{Terminology and Related Work\label{sec:prelim}}

We briefly discuss some known concepts and methods that are used in this work.

\subsection{Unsupervised Topic Modeling\label{sec:unsupervised TM}}


The model \ourmodelM introduced in this paper requires an unsupervised topic model of the dataset. To create such a model, we use Latent Dirichlet Allocation (LDA) by \citet{944937}. LDA is a statistical topic modeling algorithm that considers topics as distributions over vocabulary words, and documents as probabilistic mixtures of these topics and attempts to infer these distributions using statistical information. To model the distributions, LDA uses Dirichlet distributions and to infer their parameters, it uses a generative process whereby documents are created from words. Given a corpus $D$ and a number $K$ that determines the number of latent topics to look for in $D$, LDA attempts to assign multinomial distributions $\theta_d \sim \textit{Dir}(\alpha)$ to each document $d$, and $\phi_k \sim \textit{Dir}(\beta)$ to each topic~$k$ in such a way that the probability of $D$ is maximized if we imagine to generate each document $d$ word by word.\footnote{Here, $\alpha$ and $\beta$ are the parameters of the Dirichlet prior distributions. They reflect a-priori beliefs on the document-topic distribution and topic-word distribution, respectively.} To generate document $d$, for each position in the document, LDA first chooses a topic $k$ from $\theta_d$ and then samples a word from $\phi_k$. \citet{944937} utilize variational Bayes to estimate the optimal parameters of $\theta_d$ and $\Phi_k$.

\subsection{Semi-supervised Classification}

Semi-supervised learning is an emerging research direction attempting to find ways to deal with a lack of labeled samples by relying only on a very small gold labeled subset \Lab of the dataset. Many of the recent approaches in semi-supervised learning use consistency training on a large amount of unlabeled data~\citep{laine2016temporal,tarvainen2017mean}. These methods regularize model predictions to be invariant to small levels of noise. \citet{xie2019unsupervised} investigate the role of noise injection in consistency training and proposed Unsupervised Data Augmentation (UDA) to substitute the traditional noise injection with high quality data augmentation (e.e., back translation for textual data). 

Pseudo labeling and its extension to meta pseudo labeling are other examples of semi-supervised learning approaches. 
Pseudo labeling \citep{Lee:2013} uses two networks, called \emph{teacher} and \emph{student}. The teacher is trained on the gold-labeled portion \Lab of the dataset to predict labels, so-called pseudo labels, for the unlabeled portion \Ulab of the dataset. This turns \Ulab into the pseudo labeled dataset \Plab. $\Lab\cup\Plab$ is then used to train the student in a supervised manner. A drawback of this approach is that it lacks a mechanism for correcting inaccurate pseudo labels. To solve this problem, \citet{pham2020meta} invented the meta pseudo label approach which trains both models in an iterative fashion:
when the student has been trained with pseudo-labeled data, its performance on \Lab is used as an objective function to improve the ability of the teacher to create helpful pseudo labels.

\citet{chen2020mixtext} introduce a new text augmentation method called TMix that takes in two texts and interpolates them in their corresponding semantic hidden space. The idea behind TMix is to enforce a regularization on the model to make it behave linearly over the training data. Furthermore, the paper proposes a new semi-supervised learning method for text classification based on TMix called MixText: a text encoder (BERT) with TMix augmentation with a linear classifier on top. In each training iteration, it first predicts labels for \Ulab using the current model and then continues to train the model with \Plab using TMix.

\citet{liu-etal-2021-flitext} introduce another model, FLiText, that uses a two-stage approach, where an inspirer network based on a language model is first trained using both labeled and unlabeled data. Subsequently, this network is distilled into a smaller model. In the second stage, FLiText uses output-based distillation, which relies on the output of the inspirer, and feature-based distillation, which uses the layer weights of the inspirer to guide the training of the target network while maintaining the parameters of the inspirer network.



\citet{yang-etal-2023-prototype} introduce prototype-guided pseudo-labeling (PGPL) for semi-supervised text classification. To mitigate bias caused by imbalanced datasets, they track the number of samples used from each class in the training history. For each class, they select the $k$ nearest samples to the corresponding class prototype for the subsequent training iteration to ensure a balanced training process. Additionally, they employ prototypes for prototype-anchored contrasting, pushing samples toward their respective class prototypes and away from others.

\subsection{Masking}

Masking is a technique that can be used in training a language model (LM). It was originally applied to transformer architectures like BERT in the form of masked language modeling (MLM), to make them learn lexical and syntactic patterns from unlabeled text data. Recently, new masking tasks have been proposed to embed downstream task-related information into general pre-trained language models.
\citet{joshi2020spanbert} propose SpanBERT that masks random contiguous spans of text instead of individual tokens to better represent and predict spans of text. 

In this paper, we investigate whether pre-training based on masking can improve \ourmodel. For this, we first use LDA to find words in the dataset that carry topical information (independently of the specific topics directing the later classification). We then mask these words in a pre-training phase using the whole-word-masking approach to make the language model more sensitive to topic information in the dataset. In contrast to the work of \citet{gu-etal-2020-train}, our pre-training works in a completely unsupervised manner (using LDA).

\subsection{Topic Coherence\label{sec:coh_measure}}

Our method uses measures of topic coherence to find out which words to mask. Topic coherence measures take the~$N$ top words of a topic, compute confirmation values for individual words or subsets of words, and sum those conformation values up. \citet{roder2015exploring} proposed a unifying framework consisting of four parts to represent coherence measures. According to this framework, given a sequence of words $w_1,w_2,\dots,w_N$ characterizing a topic (ordered by importance), a topic coherence measure considers pairs $(\Wtar,\Wsupp)$ of subsets of $W=\{w_1,\dots,w_N\}$. For each of the considered pairs, a confirmation value $\kappa(\Wtar,\Wsupp)$ is computed based on word probabilities. This value is intended to reflect how well \Wsupp supports \Wtar. These values are then accumulated into a single value representing the overall topic coherence. Topic coherence measures differ in
\begin{enumerate*}[label=(\alph*)]
\item which pairs $(\Wtar,\Wsupp)$ are considered,
\item how, given certain word probabilities, $\kappa(\Wtar,\Wsupp)$ is defined, 
\item how word probabilities are calculated, and
\item how the values are accumulated.
\end{enumerate*}

For \ourmodelM, we use the two coherence measures \CUMass and \Cv by \citet{mimno2011optimizing} and \citet{roder2015exploring}, respectively. They are briefly described below, using the framework of \citep{roder2015exploring}.

\noindent \CUMass is given as follows.
\begin{itemize}
    \item The pairs considered are all $(\Wtar,\Wsupp)=(\{w_i\},\{w_j\})$ (simplified to $(w_i,w_j)$ below) with $1\le j<i\le N$.
    \item The probability of $w_i$ is defined to be the fraction of documents in which $w_i$ occurs.
    \item For every pair $(w_i,w_j)$, the confirmation value is defined to be the logarithm of the conditional probability of $w_i$ given $w_j$, slightly adjusted by a small term~$\epsilon$ to avoid taking the logarithm of zero:
    \[
        \kappa(w_i,w_j)=\log\frac{P(w_i,w_j)+\epsilon}{P(w_j)}.
    \]
    \item The overall topic coherence is the average of all $\kappa(w_i,w_j)$, $1\le i<j\le N$.
\end{itemize}

\noindent \Cv is a more complex measure given as follows:
\begin{itemize}
    \item The considered pairs are all $(\Wtar,\Wsupp)=(\{w_i\},W)$, for $1\le i\le N$,
    \item The probability of $w_i$ is defined to be the fraction of \emph{virtual} documents in which $w_i$ occurs. For this, a sliding window technique is used: viewing every original document as a string of words, every substring of length $\lambda$ is a virtual document. (We use the same $\lambda$ as Röder et al., namely $\lambda=110$.)
    \item An indirect measure is used to capture semantic relations: first, a direct measure $\kappa_0$ is used to map every word $w_i$ to a context vector $v_i=(v_{i1},\dots,v_{iN})$ by setting $v_{ij}=\kappa_0(w_i,w_j)$.\footnote{The precise confirmation measure $\kappa_0$ used by \Cv is not so important for the present discussion
    .} Moreover, $v_W=\sum_{i=1}^N v_i$. Now, the confirmation measure used by \Cv is $\kappa(w_i,W)=\mathrm{sim_{cos}}(v_i,v_W)$, where $\mathrm{sim_{cos}}$ denotes cosine similarity.
    \item Again, the overall topic coherence is the average of all $\kappa(w_i,W)$, $1\le i<j\le N$.
\end{itemize}

Since the indirect confirmation of \Cv captures even semantic relations that may not materialize as direct confirmation, \Cv is considered closer to human perception of coherence; cf.~\citep{roder2015exploring}). The downside of \Cv is a much larger running time resulting from the consideration of a large number of virtual documents and the computation of the context vectors. However, we note also that \Cv requires only $\Omega(N)$ memory cells during segmentation, whereas \CUMass requires $\Omega(N^2)$. Therefore, even though \CUMass has a shorter running time, it may reach the limit of available memory if the number $N$ of words describing each topic is large.

\section{\ourmodel and \ourmodelM\label{sec:method}}

This section describes both \ourmodel, as proposed by \citet{Hatefi.etAl:21}, and its extension to \ourmodelM, proposed in the current paper.

\ourmodel uses the iterative teacher-student architecture described in the introduction, which leverages pseudo labels created by the teacher to teach the student, improves the teacher by observing the resulting performance of the student, and iterates the process.%
\begin{figure}[p!]
	\centering
        \includegraphics[width=1.5\linewidth, angle=90]{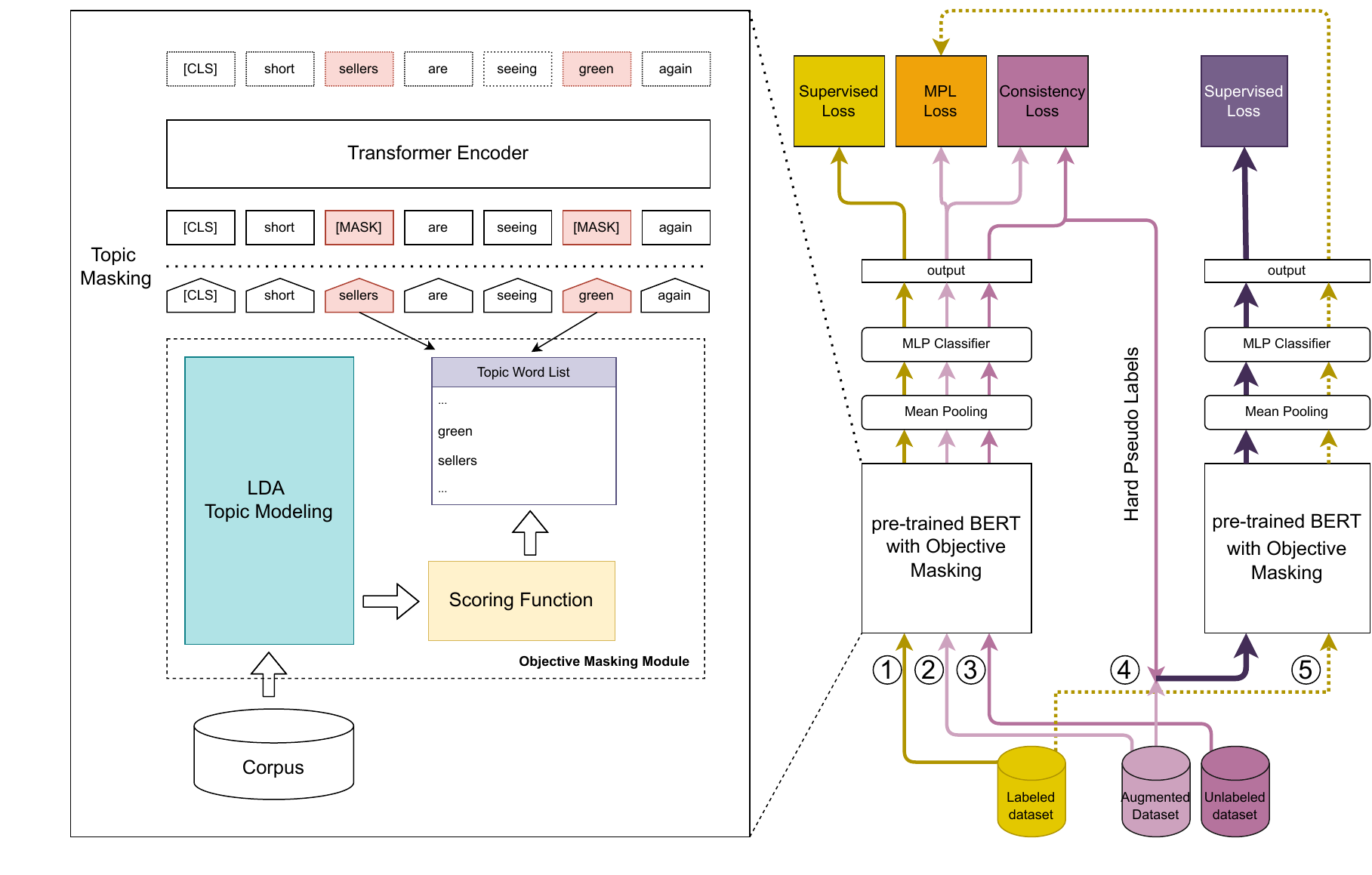}
	\caption{\ourmodelM architecture: BERT encoders are pre-trained on the dataset via Objective Masking.
	}
	\label{fig:architectureM}
\end{figure}

\subsection{\ourmodel\label{sec:base architecture}}
A schematic overview of \ourmodel can be seen in Figure \ref{fig:architectureM}. It shows the teacher~$T$ on the left and the student~$S$ on the right. The teacher is trained with the Unsupervised Data Augmentation (UDA) objective~\citep{xie2019unsupervised} and feedback consisting of the performance of the student on \Lab. Thus, the UDA objective consists of supervised loss on \Lab and consistency loss between \Ulab and an augmented version \Aug of \Ulab. The dataset \Aug can be built by applying a suitable text augmentation technique such as word substitution. In our implementation, we replace words with similar substitutes based on contextual word embeddings with probability~$0.9$.

The training data \Plab for the student is obtained from \Aug by labeling it with labels computed by the teacher, so-called pseudo labels. Thus, in a slight deviation from the earlier, somewhat simplified description, \Plab is based on \Aug rather than \Ulab for the purpose of making the student invariant to noise by regularization. The student is then trained with supervised loss on \Plab. As Figure \ref{fig:architectureM} shows, both the student and the teacher consist of an encoder that maps documents to their distributed vector representations (using a transformer and a mean pooling module that computes the average of the transformer outputs in different positions) followed by a classifier. 

To summarize, we have the following sets of training data and their derivatives:
\begin{itemize}
\item \Lab is the gold-labeled portion of the original training data. Below, we will use the notations $\lab$ and $\lbl\lab$ to denote an arbitrary element of $\Lab$ and its label, respectively.
\item $\Ulab$ is the unlabeled portion of the original training data. Elements of $\Ulab$ will be denoted by $\Ulab$.
\item $\Aug$ is the augmented version of $\Ulab$, obtained using UDA. The element of $\Aug$ that is the augmented version of $\Ulab$ will be denoted by $\aug$.
\item $\Plab$ is $\Aug$ enhanced by pseudo labels provided by the teacher. The element of \Plab obtained from $\aug\in\Aug$ will be denoted by $\plab$ and the corresponding pseudo label by $\lbl\plab$.
\end{itemize}

In each training step, a batch $\Lab'\subseteq\Lab$ of labeled data (track \mycircled{1} in Figure \ref{fig:architectureM}), and a batch $\Ulab'\subseteq\Ulab$ of unlabeled data (track \mycircled{3}) and its augmented version $\Aug'\subseteq\Aug$ (track \mycircled{2}) are fed into the teacher model. The cross entropy loss on the gold-labeled batch $\Lab'$ is computed as the mean cross entropy loss between labels $\lbl\lab$, for $\lab\in\Lab'$, and teacher predictions $T(\lab; \theta_T)$:
\[
\Loss_T(\Lab')=\sum_{\lab\in\Lab'}\frac{\mathit{Cross Entropy}(\lbl\lab, T(\lab; \theta_T))}{|\Lab'|} \enspace.
\]

Unsupervised consistency loss is computed using $\Ulab'$ and $\Aug'$. The consistency loss constrains the model predictions to be less sensitive to input noise by demanding that the model assign the same labels to augmented samples $\aug\in\Aug'$ as to the predictions of the teacher for $\Ulab'$. To encourage the model to predict confident low-entropy labels for unlabeled data, we use a sharpening function applied to the soft predictions $T(\Ulab; \theta_T)$ of the teacher for $\lbl\Ulab$. For this, we apply the same sharpening function as \citet{chen2020mixtext} to the soft predictions $T(\Ulab; \theta_T)$ of the teacher. Thus, given a temperature hyperparameter~$t$, we let
$\lbl[\shrp]\Ulab = \shrpn(T(\Ulab; \theta_T),t)$, where $\shrpn(T(\ell,t))=\frac{\sqrt[t]{\ell}}{\norm{\sqrt[t]{\ell}}}$ for a soft label $\ell$. 
Here, $\norm v$ is the $l_1$-norm of a vector~$v$ and $\sqrt[t]{\phantom X}$ is applied componentwise to a vector. Thus, without any adjustments the unsupervised loss of the teacher would be 
\[
\Loss_T(\Ulab') = \sum_{\Ulab\in\Ulab'}\frac{\mathit{Cross Entropy}(
	\lbl[\shrp]\Ulab, T(\aug; \theta_T))}{|\Ulab'|}\enspace .
\]
However, we found it beneficial to omit samples that the current model is not confident about. Therefore, the consistency loss for each batch is computed only on samples whose maximal probability over all clusters is greater than an experimentally determined confidence threshold~$\beta$. This gives rise to the unsupervised loss of the teacher which we denote by $\Loss_T(\Ulab')$.

Next, the student model learns from \Plab. For this, the soft labels are turned into hard labels. Formally, for a vector $\ell$ of soft labels, let $\hrd(\ell) = \argmax_i\,\ell_i$. Then $\Loss_S(\Plab') = $ is given by
\[
\sum_{\aug\in\Aug'}\frac{\mathit{Cross Entropy}(
	\hrd(T(\Ulab; \theta_T)), S(\aug; \theta_S))}{|\Aug'|}\enspace .
\]
Providing every $\aug\in\Aug$ with the (pseudo) label $\hrd(T(\Ulab; \theta_T))$ turns $\Aug$ into $\Plab$ (cross-point \mycircled{4} in Figure \ref{fig:architectureM}) which is fed to the student. The learning objective of the student is to minimize the cross entropy loss between the  pseudo labels and its own predictions.

To complete the current iteration of the learning process, the teacher learns from the reward signal of how well the student performs on the labeled batch $\Lab'$ (signified by the dotted line \mycircled{5} from student to teacher in Figure \ref{fig:architectureM}). Following \citet{pham2020meta}, this loss is called \emph{meta pseudo labels (MPL) loss}. Denoting the updated parameters of the student by $\theta'_{S}$, we get:
\[
	\Loss_T^{\mpl}(\Lab') = \nabla_{\theta_T}\sum_{\lab\in\Lab}\frac{\mathit{Cross Entropy}(\lbl\lab,S(\lab;\theta'_S))}{|\Lab'|}\enspace .
\]
See \citep{pham2020meta} for more details. 

Combining the three losses, we define the overall objective function of the teacher as follows.
\[
	\Loss_T= \Loss_T(\Lab') + \lambda_u * \Loss_T(\Ulab') + \Loss_T^{\mpl}(\Lab')\enspace,
\]
where $\lambda_u$ is the contribution coefficient of the consistency loss. 
To prevent overfitting to \Lab, we employ label smoothing \citep{muller2019does} when computing the supervised losses $\Loss_T$ and $\Loss_S$.

As the student only learns from pseudo-labeled data generated by the teacher, in a very final step after convergence we fine-tune it on \Lab to improve its accuracy.

\subsection{Pre-training by Objective Masking}
\label{sec:masking}
We now describe how we employ masking to attune the pre-trained language model to the dataset. For this, we create an unsupervised topic model for it, using LDA. The steps are as follows:
\begin{enumerate}
    \item We perform LDA on the given dataset to find $K$~suitable topics. 
    For each of the resulting topics $T_i$ ($1\le i\le K$), this provides us with a sorted list $L_i$ of its most indicative words.
    \item From the total vocabulary, we now choose a subset $W$ of $N$~words by selecting the most relevant words of each topic. For this we reorder $L_i$ according to a relevance measure (see below). Then if $L_i = w_0^i,w_1^i,w_2^i,\dots, w_{K-1}^i$, we let $W=\bigcup_{i=1}^K\{w_0^i,\dots,w_{j_i}^i\}$ for suitable $j_1,\dots,j_K$. (How to choose $K$ and $j_1,\dots,j_K$ is discussed below.)
    \item Afterwards, in the task-specific pre-training of the transformer model, we mask random occurrences of words on $W$ from each document in such a way that 15\% of the tokens of each document are masked. If a document does not contain sufficiently many words from $W$ to reach the 15\% limit, we mask additional random words.
    \item Finally, we use the fine-tuned language model as a basis for the teacher and student of the architecture described in Section \ref{sec:base architecture}.
\end{enumerate}

\newcommand{\lIfElse}[3]{\lIf{#1}{#2 \textbf{else}~#3}}

\begin{algorithm}[p]
\caption{\ourmodelM{}}\label{alg:cformerM}
\SetKwInOut{Input}{Input}\SetKwInOut{Output}{Output}
\small
\Input{$\Lab$, $\Ulab$ -- dataset (labeled and unlabeled)\\ 
$f_\theta$ -- pre-trained language model (BERT)\\
$W_T$ -- MLP head of teacher\\
$W_S$ -- MLP head of student\\
$K$ -- the number of topics for LDA\\
$\textit{MaxIter}$ -- maximum number of iterations\\
$B$ -- training batch size\\
$t$ -- temperature parameter for sharpening\\
$\beta$ -- \begin{tabular}{@{}l@{}}confidence threshold for\\ unlabeled dataset\end{tabular}\\
$\eta_T, \gamma_T$ -- teacher's learning rates\\
$\eta_S, \gamma_S$ -- student's learning rates\\
}
\BlankLine
\Output{%
$(\theta^{*}_T, W^*_T)$ -- learned weights of teacher \\
$(\theta^{*}_S, W^*_S)$ -- learned weights of student \\
}
\BlankLine

$\textit{TM}_K \leftarrow$ run LDA on $\Lab\cup\Ulab$ to find $K$ topics\;
$\textit{TWL} \leftarrow$ topic word list from $\textit{TM}_K$\Comment{see Section~\ref{sec:choose list}}\\
$f_{\theta_T}, f_{\theta_S} \leftarrow$ \begin{tabular}[t]{@{}l@{}}pre-train $f_\theta$ on $D$ with MLM objective\\ using $\textit{TWL}$ for masking\end{tabular}\;
\BlankLine
\BlankLine
$\Aug \leftarrow$ do augmentation for $\Ulab$\;
\For{$\textit{iter} = 1 ~\KwTo~\textit{MaxIter}$}{
$\Lab'$, $\Ulab'$ $\leftarrow$ batches of size $B$ from $\Lab$ and $\Ulab$\;
$\Aug'$ $\leftarrow$ augmented version of $\Ulab'$ from $\Aug$\;
$L_T$ $\leftarrow$ 
$W_T(f_{\theta_T}(\Ulab))$\;
\textit{// Training teacher}\;
compute $\textit{Loss}_T(\Lab')$ \Comment{teacher's supervised loss}\\
$L_T^{\textit{sharp}} \leftarrow \frac{\sqrt[t]{L_T}}{\norm{\sqrt[t]{L_T}}}$\;
compute $\textit{Loss}_T(\Ulab')$
\\\Comment{teacher's unsupervised loss}\\
$L_T^{\textit{hard}} \leftarrow \argmax_i\,L_T^i$\;
\BlankLine
\textit{// Training student}\;
$L_S \leftarrow W_S(f_{\theta_S}(\Aug'))$\;
$\textit{Loss}_S \leftarrow$ cross-entropy-loss$(L_S, L_T^{\textit{hard}})$ \\\Comment{student's supervised loss}\\
$\theta_S \leftarrow \theta_S - \eta_S*\textit{Loss}_S(\theta_S)$ \Comment{Update $\theta_S$}\\
$W_S \leftarrow W_S - \gamma_S*\textit{Loss}_S(W_S)$ \Comment{Update $W_S$}\\
\BlankLine
\textit{// Improving teacher}\;
$L_S' \leftarrow W_S(f_{\theta_S}(\Lab'))$\;
$\textit{Loss}^{\textit{MPL}}_T \leftarrow \nabla_{\theta_T}$ cross-entropy-loss$(L_S', L_{\textrm{gold}})$ \\\Comment{teacher's MPL loss}\\
$\textit{Loss}_T \leftarrow \textit{Loss}_T(\Lab') + \textit{Loss}_T(\Ulab') + \textit{Loss}^{\textit{MPL}}_T$\;

$\theta_T \leftarrow \theta_T - \eta_T*\textit{Loss}_T(\theta_T)$ \Comment{Update $\theta_T$}\\
$W_T \leftarrow W_T - \gamma_T*\textit{Loss}_T(W_T)$ \Comment{Update $W_T$}
}
\Return $(\theta^{*}_T, W^*_T)$, $(\theta^{*}_S, W^*_S)$\;

\end{algorithm}

Algorithm~\ref{alg:cformerM} illustrates the overall algorithm underlying \ourmodelM including the pre-training phase with objective masking.

\subsection{Choosing the Number of Topics for Topic Modeling\label{sec:choose number}}
For the LDA topic modeling, we need to provide the LDA algorithm with the number~$K$ of topics of the model to be created. To find a suitable $K$, we can use coherence measures such as \CUMass and \Cv. Here, we apply \Cv. Thus, given a dataset, we run the LDA algorithm on it for a range of candidate values for~$K$ and compare the resulting topic models with respect to \Cv. We determine~$K$ from the coherence plot using the well-known heuristics of the ``elbow method''~(cf.~\citet{Blashfield.etAl:82}). This rule helps identify the point where the rate of increase in the coherence scores starts to level off, resulting in an ``elbow'' shape in the plot. In the degenerate case that the first point of the graph is the highest one, we choose that one for our experiments.

\subsection{Choosing Word Lists for Masking\label{sec:choose list}}

After choosing the most promising topic model for the dataset, we extract the $N$ most relevant words for each topic and compile them into a list. The selection of $N$ is guided by heuristics, and we subsequently assess the quality of each list to identify the best one. The list quality is evaluated by computing the average coherence of all topics, considering their respective $N$ most relevant words. To compute the topic coherence we use the coherence measures explained in Section \ref{sec:coh_measure}.

For a given $N$, the actual selection of the $N$ most relevant words from each topic uses the relevance measure introduced by \citet{sievert2014ldavis}. Let $\phi_{kw}$ denote the probability of a word $w$ to occur in a document of topic $k \in \{1, \dots, K\}$, and let $p_w$ denote the marginal probability of $w$ in the entire corpus. The \emph{relevance} of word $w$ to topic $k$ given a weight parameter $\lambda$ (where $0\leq \lambda \leq 1$) is defined as:
\[
r(w,k|\lambda) = \lambda\log(\phi_{kw}) + (1-\lambda)\log(\frac{\phi_{kw}}{p_w})\enspace.
\]
The $N$ words picked from topic $k\in\{1,\dots,K\}$ are the $N$ most relevant words of topic~$k$ according to this measure.
When $\lambda$ is small, words that are highly associated with the topic and not very common in other topics will receive higher relevance scores. This makes the topics more distinct and easily interpretable. However, it might not consider words that are relevant but are more common across multiple topics. Conversely, when $\lambda$ is higher, words that are more frequent in the topic but also have higher general frequency across topics will receive higher relevance scores. The advantage of using a larger $\lambda$ is that it can capture more general aspects of the topic, helping to identify commonly occurring terms related to the topic across different documents.

\section{Experiments and Analysis}
\label{sec:exp_analyse}
Our experiments revolve around answering the following research questions: 
\begin{enumerate}[label={\bfseries\mathversion{bold}\question{\arabic*}:}]
    \item How should one choose the parameters of the topic word selection method?
    \item What is the overall performance of \ourmodel in comparison to the baselines?
    \item How is the overall performance of \ourmodel and \ourmodelM affected if the BERT model of the student is replaced by DistilBERT, yielding Distil-\ourmodel and Distil-\ourmodelM, respectively?
    \item Does the objective masking used in \ourmodelM indeed improve \ourmodel?
    \item What factors could impact the effectiveness of the proposed objective masking?
    \item Are there any benefits to utilizing topic modeling for generating topic word lists as opposed to simpler methods such as TF-IDF?
    \item Does the proposed masking approach affect the reliability and interpretability of \ourmodel?
    \item How does the performance of the models vary with different values of hyper-parameters such as the number of GPUs and the batch size? 
    \item How well does the proposed model perform in a zero-shot setting in comparison to the baselines?
\end{enumerate}

\subsection{Experimental Setup}
\subsubsection*{Datasets}
We perform experiments with two English text classification benchmark datasets and a private dataset which is in Swedish. The dataset statistics and splits are given in Table \ref{tab:data_set}.

\begin{table}[ht!]
\resizebox{\textwidth}{!}{%
\begin{tabular}{@{}c@{\hspace{1.5ex}}c@{\hspace{1.5ex}}c@{\hspace{1.5ex}}c@{\hspace{1.5ex}}c@{\hspace{1.5ex}}c@{\hspace{1.5ex}}c@{\hspace{1.5ex}}c@{\hspace{1.5ex}}c@{\hspace{1.5ex}}c@{\hspace{1.5ex}}c@{}}
\toprule
\textbf{Dataset} &
  \textbf{Classes} &
  \textbf{Documents} &
  \textbf{Average \#s} &
  \textbf{Max \#s} &
  \textbf{Average \#w} &
  \textbf{Max \#w} &
  \textbf{Vocabulary} &
  \textbf{Unlabeled} &
  \textbf{Dev} &
  \textbf{Test} \\ \midrule
  \yahoo &
  10 &
  140\,000 &
  7.0 &
  158 &
  112.0 &
  2\,001 &
  267\,610 &
  5\,000 &
  5\,000 &
  6\,000 \\
\ag &
  4 &
  120\,000 &
  1.7 &
  20 &
  36.2 &
  212 &
  94,443 &
  5\,000 &
  2\,000 &
  1\,900 \\
\bonnier &
  17 &
  78\,757 &
  22.23 &
  382 &
  362.12 &
  2\,252 &
  538\,655 &
  -- &
  -- &
  -- \\\bottomrule
\end{tabular}%
}
\caption{Dataset statistics and dataset split. The numbers of sentences and words are denoted by \#s and \#w, respectively. The number of unlabeled, dev, and test data items is given in terms of the number of data items per class. For \yahoo, the reported statistics concerns the subset used in the experiments.\label{tab:data_set}}
\end{table}

\begin{description}
\item[\yahoo] For \yahoo~\citep{chang2008importance}, we obtain the text to be classified by concatenating the question (title and content) and the best answer. Since the original training set is very large, we only use a randomly chosen subset of 10\% of its documents (i.e.,~140\,000 documents). To be comparable with our baselines, we randomly sample the same amount of data as in MixText~\citep{chen2020mixtext} from the training subset for our unlabeled and validation sets, and use the original \yahoo test set. We use the training subset to create word lists for objective masking.

\item[\ag] Of \ag~\citep{zhang2015character}, we only use the news content (without titles). Again, we randomly sample the same amount of data as in MixText~\citep{chen2020mixtext} from the original training set for our unlabeled and validation sets and use the original \ag test set. We use the original training set to create word lists for objective masking.

\item[\bonnier] The \bonnier\footnote{\url{https://www.bonniernews.se/}; while we cannot make this dataset publicly available, researchers who want to reproduce our results or use it for their own work may contact \texttt{\href{mailto:datasets@bonniernerws.se}{datasets@bonniernerws.se}} to gain access.} is a private dataset comprised of 127\,161 articles in Swedish published on 35 different Bonnier News brands during the period February 2020 to February 2021. Its documents are labeled according to the Category Tree for Swedish Local News that has been developed and used by local newsrooms within Bonnier News\footnote{\url{github.com/mittmedia/swedish-local-news-categories}}. This category tree is based on the IPTC Media Topics\footnote{\url{iptc.org/standards/media-topics/} is a comprehensive standard taxonomy for categorizing news text.}.
The dataset includes 545 categories distributed across four hierarchy levels
. The dataset is highly imbalanced, with the most frequent category occurring 30\,531 times and the least frequent one occurring 102 times. Furthermore, the number of categories articles are labeled with varies greatly. The maximum number of categories used to label an article is 46 and the minimum number is one, with 5.1 categories on average. For our experiments, we only consider top-level labels and use only documents labeled with a unique label. The resulting dataset consists of 78\,757 samples in 17 classes. We split these samples into test and training datasets according to a 1:4 ratio and randomly select 20\% of the training examples for validation. Since the dataset is imbalanced, instead of choosing an absolute number of examples per class to create the labeled and unlabeled datasets, we split samples of each class in the training data into labeled and unlabeled parts using 1\%, 10\%, and 30\% as labeled documents, respectively. Table~\ref{tab:bonnier_stat} displays the classes in the dataset along with the corresponding number of examples in the training, validation, and test sets. 
\end{description}

\begin{table}[ht!]
\newcommand{\two}[1]{\begin{tabular}{@{}ll@{}}#1\end{tabular}}
\centering\small
\begin{tabularx}{0.9\linewidth}{@{}X@{}rrr@{}}
\toprule
\textbf{Class name}   &\textbf{Training} & \textbf{Test} & \textbf{Valid.} \\\midrule
\two{Olyckor \& katastrofer\\(Accidents \& disasters)}                & 2656 & 830  & 664  \\\midrule
\two{Brott \& straff\\(Crime \& punishment)}           & 6587     & 2059 & 1647       \\\midrule
Personligt (Personal)                           & 1503     & 470  & 376        \\\midrule
\two{Vetenskap \& teknologi\\(Science \& technology)}  & 235      & 74   & 59         \\\midrule
\two{Samhälle \& välfärd\\(Society \&   welfare)}      & 3198     & 1000 & 800        \\\midrule
Religion \& tro (Religion \& faith)             & 185      & 57   & 46         \\\midrule
\two{Ekonomi, näringsliv   \& finans\\(Economy, business   \& finance)}  & 5222 & 1632 & 1305 \\\midrule
Politik (Politics)                           & 2423     & 757  & 606        \\\midrule
Sport (Sports)                               & 16746    & 5233 & 4187       \\\midrule
\two{Livsstil \& fritid\\(Lifestyle \& leisure)}    & 1922     & 601  & 480        \\\midrule
Miljö (Environment)                          & 911      & 285  & 228        \\\midrule
Väder (Weather)                              & 454      & 142  & 114        \\\midrule
\two{Hälsa \& sjukvård\\(Health \& medical care)}   & 1886     & 589  & 471        \\\midrule
\two{Konflikter, krig \&   terrorism\\(Conflicts, war \& terrorism)}  & 70   & 21   & 17   \\\midrule
\two{Kultur \& nöje\\(Culture  \&  entertainment)}                    & 4829 & 1509 & 1207 \\\midrule
\two{Skola \& utbildning\\(School \& education)}    & 1201     & 375  & 300        \\\midrule
Arbetsmarknad (Labor market)                 & 376      & 118  & 94   \\\bottomrule     
\end{tabularx}%
\caption{Classes and sample distribution in \bonnier across Training, Validation, and Test sections\label{tab:bonnier_stat}}
\end{table}

\subsubsection*{Baselines and Experimental Settings}

To verify the effectiveness of \ourmodel and \ourmodelM, we compare them with several baselines:
\begin{description}
\item[UDA~\citep{xie2019unsupervised}] uses the consistency loss between unlabeled and augmented data as a training signal to improve classification. 
\item[MixText~\citep{chen2020mixtext}] augments training samples by interpolating in the hidden space. 
\item[FLiText~\citep{liu-etal-2021-flitext}] applies pseudo-labeling and distillation to a lightweight setting using convolution networks. 
\item[PGPL~\citep{yang-etal-2023-prototype}] proposes prototype-guided pseudo-labeling to avoid bias from imbalanced data and presents prototype-anchored contrasting to make clear boundaries between classes.
\item[BERT/DistilBERT] consists of a BERT/DistilBERT encoder followed by a two-layer MLP serving as the classifier, similar to the \ourmodel classification layer. This classifier is exclusively trained with labeled data. Moreover, we have pre-trained versions of the BERT/DistilBERT classifier, utilizing pre-trained BERT models tailored for use in the \ourmodelM classifier.
\end{description}

We implement MixText and UDA using the code\footnote{\url{https://github.com/SALT-NLP/MixText}} provided by \citet{chen2020mixtext} with the hyperparameters specified in the original paper and the code repository. Similarly, for FLiText, we run the available code\footnote{\url{https://github.com/valuesimplex/FLiText}} on \ag and \yahoo, using the hyperparameters reported in the original paper and its code repository. We implement the BERT/DistilBERT baselines ourselves. 

For all baselines, we report the average results obtained from five different runs, just as we do for different versions of the \ourmodel model. It is worth noting that all baseline models are trained using the same amount of labeled, unlabeled, and validation data as used in our \ourmodel experiments. Additionally, we apply the same text augmentation approach whenever augmented data is needed for the baseline models.

Our experiments evaluate two versions of \ourmodel and their \ourmodelM counterparts:
\begin{description}
	\item[\ourmodel:] The student and the teacher models both employ the BERT language model. 
	\item[Distil-\ourmodel:] The teacher is as in \ourmodel but the student uses DistilBERT as its language model.
	\item[\ourmodelM and Distil-\ourmodelM:] These are the \ourmodelM versions of the two previous models.
\end{description}

For the English datasets, we use \emph{bert-base-uncased} and \emph{distilbert-base-uncased} as the BERT and DistilBERT language models, respectively. For Swedish, we use the pre-trained BERT model available in the hugging face repository\footnote{\url{https://huggingface.co/KB/bert-base-swedish-cased}},
namely \emph{KB/bert-base-swedish-cased} as the BERT language model and  \emph{distilbert-base-multilingual-cased} as the DistilBERT language model. The Swedish models are referred to as \ourmodel(SE-SE) and Distil-\ourmodel(SE-multi) to indicate the types of language model used by teachers and students, respectively. To have a fair comparison between \ourmodel and Distil-\ourmodel, we also train a \ourmodel called \ourmodel(SE-multi) with the Swedish BERT as the teacher's encoder and the \emph{bert-base-multilingual-cased} model for the student's encoder.
The teacher and student models in the English \ourmodel have 109.58 million parameters each, and the student in the English Distil-\ourmodel has 66.46 million parameters. In the Swedish version of \ourmodel, the teacher model has 124.79 million parameters and the student model has 124.79 and 177.95 million parameters when we use \emph{KB/bert-base-swedish-cased} and \emph{bert-base-multilingual-cased}, respectively, as its language model. The student in the Swedish Distil-\ourmodel has 134.83 million parameters.

We employ average pooling over the output of the encoder to aggregate word embeddings into document embeddings, and a two-layer MLP with a 128 hidden size and hyperbolic tangent as its activation function (the same as in MixText) to predict the labels. For the input of the model, documents are truncated to their first 256 tokens.
To generate augmented data from unlabeled data, we use the library \emph{nlpaug}\footnote{\url{https://github.com/makcedward/nlpaug}}. We substitute text words based on contextual word embeddings with a probability of 0.9. All models are trained with the AdamW optimizer~\citep{loshchilov2017decoupled}.
We train the models for 7\,000 steps (including 50 warm-up steps) and evaluate them every 500 steps. To avoid overfitting, we use early stopping with delta 0.005 and patience 4. We set the learning rate of the transformer and classifier components in both teacher and student to 1e-5 and 1e-3, respectively.
After training both the teacher and the student models, we fine-tune the student on the labeled data set using the AdamW optimizer with a fixed learning rate of 5e-6 and a batch size of 32, running for 10 epochs. The temperature $T$ for sharpening is set to 0.5 for \yahoo and \bonnier and to 0.3 for \ag. The confidence threshold $\beta$ is set to 0.9 and the label smoothing parameter is 0.15 for all datasets. For the contribution coefficient of unsupervised loss in the teacher loss function $\lambda_u $, we start from 0 and increase it linearly for 6\,000 steps until it reaches its highest value of 1.

We use the PyTorch Distributed package for distributed GPU training on 3 V100 GPUs with 32GB memory. However, it is not mandatory to train \ourmodel on such large memory GPUs. With small batch sizes, the model can be trained using regular GPUs.
Table \ref{tab:gpu_batch} shows the number of GPUs and local batch sizes\footnote{The local batch size is the batch size per GPU.} we used for training the models across three datasets.

\begin{table*}
\resizebox{\textwidth}{!}{%
\begin{tabular}{lcccc}
\toprule
Dataset        & \begin{tabular}{@{}c@{}}Number\\ of GPUs\end{tabular} & \begin{tabular}{@{}c@{}}Labeled Batch Size\\ per GPU\end{tabular} & \begin{tabular}{@{}c@{}}Unlabeled Batch Size\\per GPU\end{tabular} & \begin{tabular}{@{}c@{}}Occupied Memory\\per GPU\end{tabular} \\\midrule
\ag        & 3     & 4                      & 8                        & 15.4GB               \\
\yahoo & 2     & 8                      & 16                       & 23.6GB                   \\
\bonnier   & 2     & 4                      & 8                        & 15.4GB\\
\bottomrule 
\end{tabular}%
}
\caption{The number of GPUs and local batch sizes across three datasets.\label{tab:gpu_batch}}
\end{table*}

\subsection{Choosing Suitable Parameters for the Selection of Topic Words (\question1)\label{sse:question 1}}
This section reports on our experiments with the topic word list selection method to determine the appropriate values for its parameters across different datasets.

\paragraph{Choosing the Number of Topics using LDA}
To create the LDA topic model as described in Section \ref{sec:choose number}, we use the Gensim library\footnote{\url{https://radimrehurek.com/gensim/}}. Before feeding the documents into the model, a preprocessing step is performed, involving the elimination of stop words, removal of some of the most frequent words in the dataset, and retention of only nouns, adjectives, and verbs. For \bonnier, we also perform lemmatization using the spaCy library\footnote{\url{https://spacy.io/models/sv}}. Swedish has considerably more inflections than English. For example, the definite article in Swedish is mostly expressed by a suffix on the noun, and agreement rules stipulate that adjectives are inflected depending on the gender and number of the nouns they refer to, for instance: \emph{en fin bil} (a beautiful car), \emph{ett fint hus} (a beautiful house), and \emph{fina bilar} (beautiful cars). Hence, performing lemmatization during preprocessing is justified for Swedish.  

\begin{figure}[ht!]
    \centering
    \includegraphics[width=0.8\linewidth]{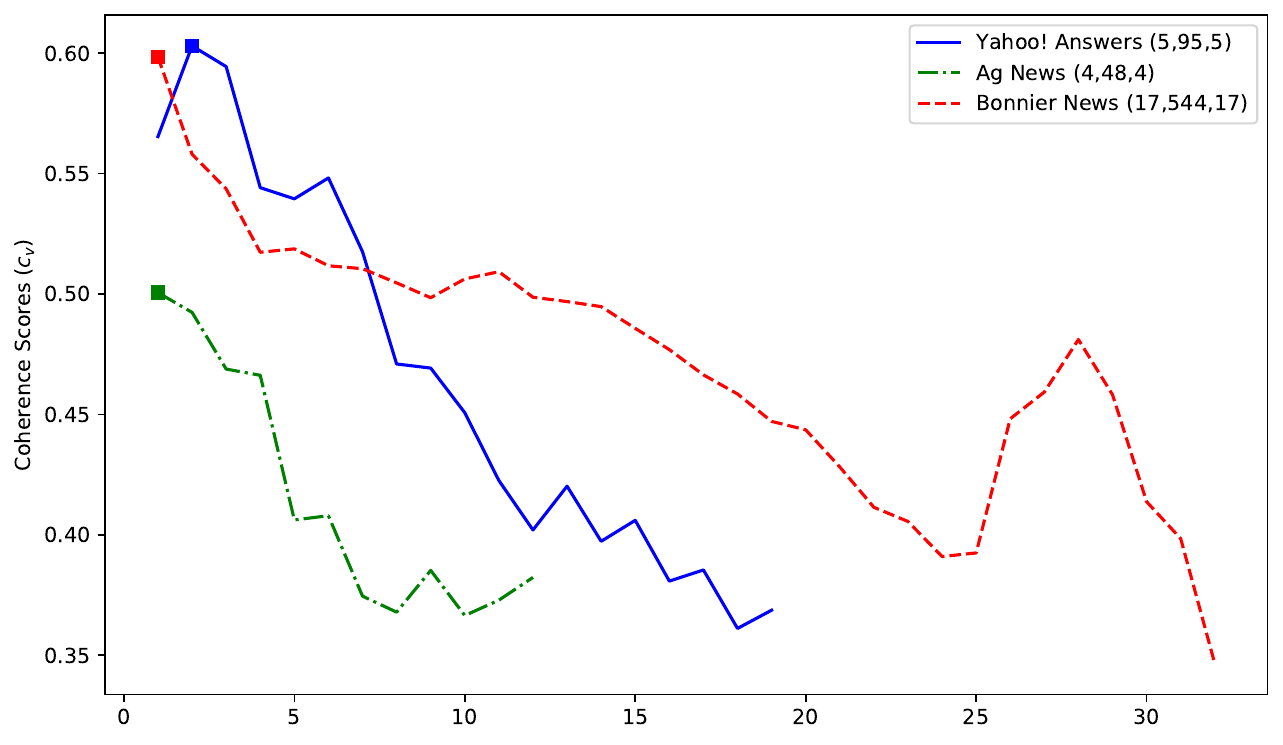}
    \caption{The coherence diagram over a range of values of the number of topics for 
(a) \yahoo (b) \ag, (c) \bonnier. Note that values on the $y$-axis range between $0.35$ and $0.6$. For each dataset, the range of values explored (resulting from the respective choice of $m$ and $k$) is denoted within parentheses in the legend. For instance, ``\yahoo (5, 95, 5)'' means that the number of topics ranges from 5 to 95 with a step size of 5, in this case, $m=5$ and $k=19$.}
\label{fig:topic_coh}
\end{figure}

Now, our goal is to determine the number of topics that maximizes the coherence score of the resulting LDA model. To determine this number, we used the following reasoning. As we work in a semi-supervised setting, we know a lower bound on the number of topics in each dataset, namely the number of distinct labels occurring in the labeled portion of the dataset. If this number is $m$, we compute coherence scores of $k$ LDA models, each dividing the dataset into $im$ topics for $i=1,\dots,k$. The number $k$ should be sufficiently large to ensure that the highest coherence score is likely to be included. For the three datasets considered in this paper, we use $k=19$ (\yahoo), $k=12$ (\ag), and $k=32$. We then choose $i_{\max} k$ as the number of topics for the LDA model to be used, where $i_{\max}$ is the value of $i$ resulting in the highest coherence score. Figure \ref{fig:topic_coh} shows the resulting coherence diagrams for the three datasets. For \yahoo, we actually chose to set $m=5$ rather than $m=10$ in order to confirm that smaller numbers of topics than the actual number would not result in higher coherence scores. Figure \ref{fig:topic_coh} shows the coherence diagrams for the three datasets. As one may expect, the result would have been  $i_{\max}=1$ in all three cases if we had chosen $m=10$ in the case of \yahoo. Based on the curves shown in Figure \ref{fig:topic_coh}, we choose 10 topics for \yahoo, 4 for \ag, and 17 for \bonnier as presumably good numbers of topics for our LDA models.

These experimental results indicate that it may usually be safe to skip the coherence calculations, instead simply choosing the number of topics that coincides with the number $m$ of distinct labels found in the labeled portion of the data. We thus recommend to do this unless there is reason to suspect that the labeled portion of the data does not cover all of the classes.

\paragraph{Relevance-Based Word Lists} For each dataset, we choose relevance-based word lists for masking as explained in Section \ref{sec:choose list}. Thus, we create several candidate lists with varying values of $N$, comprising the most informative words from the vocabulary. Table \ref{tab:list_coh} presents the coherence scores ($C_v$ and \CUMass) of these lists for different datasets. For each dataset, we create two sets of lists, one corresponding to $\lambda = 0.7$ and the other to a lower $\lambda$. From each set, we select one list. The choice of the lower $\lambda$ value was made based on dataset examination, aided by topic model visualization tools\footnote{\url{https://pypi.org/project/pyLDAvis/}}.

\begin{table}[t!]
\centering
\begin{tabularx}{0.7\linewidth}{Xcccc}
\toprule
Dataset & $\lambda$ & $N$ & $C_v$ & \CUMass \\\midrule
\multirow{10}{*}{\begin{tabular}[c]{@{}l@{}}\yahoo\end{tabular}}    & 0.2   &  500  &  0.68  &  -13.34  \\
 &  \textbf{0.2}  &  \textbf{1000}  &  \textbf{0.79}  &  \textbf{-15.75}  \\
 &  0.2  &  1500  &  0.80  &  -15.55 \\
 &  0.2  &  2000  &  0.81  &  -15.46  \\\cmidrule(lr){2-5}
 &  0.7  &  500   &  0.54  &  -9.11 \\
 &  0.7  &  1000  &  0.65  &  -11.95  \\
 &  \textbf{0.7}  &  \textbf{1500}  &  \textbf{0.72}  &  \textbf{-13.48} \\
 &  0.7  &  2000  &  0.77  &  -14.50  \\
 &  0.7  &  2500  &  0.80  &  -15.56  \\
 &  0.7  &  3000  &  0.83  &  -16.23  \\
\midrule
\multirow{11}{*}{\begin{tabular}[c]{@{}l@{}}\ag\end{tabular}}  &  0.2  &  500  &  0.61  &  -13.78 \\
 &  0.2  &  700  &  0.69  &  -15.31  \\
 &  \textbf{0.2}  &  \textbf{900}  &  \textbf{0.74}  &  \textbf{-16.31}  \\
 &  0.2 &  1100  &  0.77  &  -16.94 \\
 &  0.2 &  1300  &  0.80  &  -17.44  \\ \cmidrule(lr){2-5}
 &  0.7  &  500  &  0.47  &  -10.03 \\
 &  0.7 &  1000  &  0.63  &  -13.30 \\
 &  0.7 &  1500  &  0.72  &  -15.32 \\
 &  \textbf{0.7} &  \textbf{2000}  &  \textbf{0.78}  &  \textbf{-16.58} \\
 &  0.7 &  2500  &  0.82  &  -17.43 \\
 &  0.7 &  3000  &  0.85  &  -18.00 \\
\midrule
\multirow{9}{*}{\begin{tabular}[c]{@{}l@{}}\bonnier\end{tabular}} &  0.1  &  500  &  0.79  &  -16.84  \\
 &  \textbf{0.1}  &  \textbf{1000}  &  \textbf{0.83}  &  \textbf{-17.19}  \\
 &  0.1  &  1500  &  0.83  &  -16.96 \\
 &  0.1  &  2000  &  0.83  &  -16.63  \\\cmidrule(lr){2-5}
 &  0.7  &  500   &  0.54  &  -9.70 \\
 &  0.7  &  1000  &  0.61  &  -11.58 \\
 &  \textbf{0.7}  &  \textbf{1500}  &  \textbf{0.68}  &  \textbf{-12.95} \\
 &  0.7  &  2000  &  0.73  &  -13.83 \\
 &  0.7  &  2500  &  0.76  &  -14.52 \\
 \bottomrule
\end{tabularx}%
\caption{Coherence scores ($C_v$ and \CUMass) of different candidate lists corresponding to different numbers of words per topic ($N$) for \ag, \yahoo, and \bonnier. $\lambda$ is the parameter of the relevance score used for choosing the $N$ most relevant words of the topics. Selected values are emphasized in boldface letters.\label{tab:list_coh}}
\end{table}

A larger value of $C_v$ and a higher absolute value of \CUMass indicate stronger coherence.
To determine the appropriate value of $N$, we identify the point that marks the end of the rapid growth of topic coherence. The rationale behind this approach is to select a point after which the marginal increase in coherence score does not justify further enlargement of $N$. If we were to pick a larger $N$ beyond this point, less significant words would be added to the list, diluting the valuable information and leading to a reduction in the objective masking capacity.

To investigate the effectiveness of relevance-based word lists for objective masking and explore the relationship between coherence scores and classifier performance, we pre-train the BERT model on \yahoo employing topic word lists corresponding to $\lambda=0.2$ and different values of $N$ for objective masking and then compute the accuracy of the resulting BERT classifier for three semi-supervised cases (10, 200, 2\,500 labeled example per cluster). Table \ref{tab:compare_p_for_yahoo} illustrates the results of these experiments. From the table, we see that the BERT classifier performs the best when $N = 1000$. In addition, the results suggest that when we increase $N$, the model accuracy starts increasing, reaches a peak, and then starts dropping again. This seems to be a reasonable behavior: by increasing $N$, we first add informative words to the list (considering that in each topic, words are sorted with respect to relevance), but eventually, less significant words are also included, thus diluting the useful information.

\begin{table}[ht!]
\centering
\begin{tabularx}{\linewidth}{Xcccc}
\toprule
                                &                    & \multicolumn{3}{c}{\begin{tabular}{@{}c@{}}Labeled examples\\per class\end{tabular}} \\\cmidrule(lr{0pt}){3-5}
Dataset                         & $N$              & 10           & 20     & 2\,500       \\\midrule
\multirow{4}{*}{\yahoo}         & 500              & 62.4         & 71.4   & 74.3       \\
                                & \textbf{1000}             & \textbf{62.9}         & \textbf{71.5}   & \textbf{74.4}       \\
                                & 1500             & 62.6         & 71.5   & 74.4       \\
                                & 2000             & 62.4         & 71.4   & 74.3   \\   
\bottomrule
\end{tabularx}%
\caption{Performance of the BERT classifier on \yahoo using relevance-based word lists with a $\lambda$ value of 0.2 and varying values of $N$ for pre-training BERT via objective masking.\label{tab:compare_p_for_yahoo}}
\end{table}

The results of \ourmodelM and Distil-\ourmodelM using relevance-based word lists for objective masking for \yahoo and \ag are available in Table \ref{tab:all_results}. Also, the results of \ourmodelM using relevance-based word lists for objective masking for \bonnier are reported in Table \ref{tab:all_results_bonnier}.

\begin{table*}
\resizebox{\textwidth}{!}{
\begin{tabular}{@{}l@{\hspace{1.5ex}}lcccl@{\hspace{1.5ex}}lccc@{}}
\toprule
 &
   &
  \multicolumn{3}{l}{\begin{tabular}{@{}c@{}}Labeled examples\\per class\end{tabular}} &
   &
   &
  \multicolumn{3}{l}{\begin{tabular}{@{}c@{}}Labeled examples\\per class\end{tabular}} \\\cmidrule(lr){3-5}\cmidrule(lr{0pt}){8-10}
 &
  Model &
  10 &
  200 &
  2500 &
   &
  Model &
  10 &
  200 &
  2500 \\\midrule
\multirow{20}{*}{\rotatebox{90}{\yahoo}} &
  BERT &
  60.2 &
  69.6 &
  73.5 &
  \multirow{20}{*}{\hspace*{1ex}\rotatebox{90}{\ag}} &
  BERT &
  81.9 &
  88.6 &
  91.5 \\
 &
  BERT (random) &
  61.6 &
  71.3 &
  74.2 &
&
  BERT (random) &
  83.8 &
  89.0 &
  91.7 \\
&
  BERT (relevance-0.2) &
  62.9 &
  71.5 &
  74.4 &
   &
  BERT (relevance-0.2) &
  84.5 &
  89.2 &
  91.9 \\
&
  BERT (relevance-0.7) &
  62.7 &
  71.5 &
  74.3 &
   &
  BERT (relevance-0.7) &
  84.3 &
  89.1 &
  91.9 \\
&
  DistilBERT &
  60.4 &
  69.9 &
  73.2 &
   &
  DistilBERT &
  83.4 &
  88.4 &
  91.1 \\
&
  DistilBERT (random) &
  61.9 &
  71.0 &
  73.7 &
   &
  DistilBERT (random) &
  84.3 &
  89.0 &
  91.5 \\
&
  DistilBERT (relevance-0.2) &
  63.3  &
  71.5 &
  73.9 &
   &
  DistilBERT (relevance-0.2) &
  84.6 &
  89.3 &
  91.6 \\
&
  DistilBERT (relevance-0.7) &
  63.1 &
  71.4 &
  73.9 &
   &
  DistilBERT (relevance-0.7) &
  84.4 &
  89.2  &
  91.6  \\
&
  UDA$^{\spadesuit}$ (2019)&
  61.7 &
  69.7 &
  73.5 &
   &
  UDA$^{\spadesuit}$ (2019) &
  86.3 &
  89.0 &
  91.5 \\
 &
  MixText$^{\spadesuit}$ (2020) &
  64.1 &
  70,6 &
  73.7 &
   &
  MixText$^{\spadesuit}$ (2020) &
  86.9 &
  88.9 &
  91.3\\
   &
  FLiText$^{\spadesuit}$ (2021)&
  45.9 &
  65.5 &
  68.5 &
   &
  FLiText$^{\spadesuit}$ (2021) &
  77.9 &
  87.6 &
  89.1 \\
     &
  PGPL (2023) &
  \textbf{67.4} &
  70.7 &
  $-$ &
   &
  PGPL (2023) &
  87.8 &
  89.2 &
  $-$
  \\\cmidrule(lr){2-5}\cmidrule(lr){7-10}
 &
  \ourmodel &
  64.6 &
  71.9 &
  74.7 &
   &
  \ourmodel &
  88.1 &
  90.0 &
  91.9 \\
 &
  \ourmodelM (random) &
  65.1 &
  72.7 &
  75.0 &
   &
  \ourmodelM (random) &
  88.0 &
  90.1 &
  \textbf{92.2} \\
&
  \ourmodelM (relevance-0.2) &
  66.3 &
  \textbf{72.9} &
  \textbf{75.1} &
   &
  \ourmodelM (relevance-0.2) &
  88.4 &
  90.1 &
  \textbf{92.2} \\
 &
  \ourmodelM (relevance-0.7) &
  66.1 &
  72.8 &
  75.0 &
   &
  \ourmodelM (relevance-0.7) &
  \textbf{88.5} &
  \textbf{90.2} &
  \textbf{92.2} \\\cmidrule(lr){2-5}\cmidrule(lr){7-10}
 &
  Distil-\ourmodel &
  64.2 &
  71.7 &
  74.5 &
   &
  Distil-\ourmodel &
  87.2 &
  90.0 &
  91.8 \\
 &
  Distil-\ourmodelM (random) &
  64.5 &
  72.5 &
  74.7 &
   &
  Distil-\ourmodelM (random) &
  87.5 &
  89.9 &
  92.0 \\
 &
  Distil-\ourmodelM (relevance-0.2) &
   65.3  &
   72.8  &
   75.0  &
   &
  Distil-\ourmodelM (relevance-0.2) &
   88.2  &
   90.1  &
   92.1  \\
 &
  Distil-\ourmodelM (relevance-0.7) &
   65.0  &
   72.7  &
   74.9 &
   &
  Distil-\ourmodelM (relevance-0.7) &
   88.3  &
   90.1  &
   \textbf{92.2}   \\
  \bottomrule
\end{tabular}
}
\caption{Comparison of the test accuracy of \ourmodel and \ourmodelM with the baselines on \yahoo and \ag. The results are the average accuracy of 5 different runs with different random seeds. $^{\spadesuit}$ means ``run by us''.\label{tab:all_results}}
\end{table*}

\begin{table*}
\begin{tabularx}{\linewidth}{@{}llYYY@{}}
\toprule
& 
& 
\multicolumn{3}{@{}c@{}}{\begin{tabular}{@{}c@{}}Proportion of\\ labeled examples per class\end{tabular}} \\\cmidrule(l{0pt}r{0pt}){3-5}
Dataset &   Model   &   0.01    &   0.1    &   0.3\\\midrule
\multirow{11}{*}{\bonnier} & BERT-SE     &  81.7     &  85.4     &  86.7  \\
& BERT-SE (random)    &  82.1     &  85.6     &  87.0  \\
& BERT-SE  (relevance-0.1)   &  82.2     &  85.8     &  87.2  \\
& BERT-SE  (relevance-0.7)   &  82.0     &  85.7    &  87.1  \\
& MixText$^{\spadesuit}$(2020)     & 82.5    &   85.7    & 87.0 \\\cmidrule(lr{0pt}){2-5}
& \ourmodel (SE-SE)              &  83.3        & 86.5       & 87.5        \\
& \ourmodel (SE-multi)           &  81.8        & 84.8       & 85.8         \\
& \ourmodelM (SE-SE) (random)    &  \textbf{83.8}   &     86.6   &      87.6  \\
& \ourmodelM (SE-SE) (relevance-0.1) &  \textbf{83.8}  &    \textbf{86.7}    &   \textbf{87.7}   \\
& \ourmodelM (SE-SE) (relevance-0.7)  &  83.6   &     86.6   &   87.6 \\
\cmidrule(lr{0pt}){2-5}
& Distil-\ourmodel (SE-multi)   & 81.4      & 84.0     & 84.8  \\       
\bottomrule 
\end{tabularx}
\caption{Comparison of the test accuracy of \ourmodel and \ourmodelM on \bonnier with baselines. The results are the average accuracy of 5 different runs with different random seeds.
Since the purpose of Distil-\ourmodel is to have a model for limited environments and Distil-\ourmodel(SE-multi) is larger than \ourmodel(SE-SE) (and still cannot outperform it), Distil-\ourmodelM(SE-multi) is not selected for further experiments. $^{\spadesuit}$ means ``run by us''.\label{tab:all_results_bonnier}}
\end{table*}

\subsection{Result Analysis (\question2 \& \question3 \& \question4)\label{sec:result_analysis}}
Table \ref{tab:all_results} and Table \ref{tab:all_results_bonnier} present the performance of different versions of \ourmodel and \ourmodelM in comparison to each other and against baseline models on three datasets.

\paragraph*{Overall performance of \ourmodel}
Overall, \ourmodel performs better than current SoTA models across all considered datasets. \ourmodel consistently exhibits significant performance improvements over UDA, ranging from 0.4\% to 2.9\% across all experiments conducted on \ag and \yahoo. A noteworthy aspect is that the teacher in \ourmodel is trained with the UDA objective. This difference in performance strongly suggests that the knowledge distillation from student to teacher within the teacher-student architecture does indeed have a significant effect. Moreover, across all experiments and datasets, \ourmodel demonstrates superior accuracy compared to both MixText and BERT/DistilBERT. It outperforms FLiText by a significant margin as well, which may be thanks to the larger size of the student model compared to FLiText's target network which is designed to be lightweight. Last but not least, \ourmodel outperforms PGPL on \yahoo and \ag in all cases except the 10-shot case of \yahoo. We note here that there is no code available for the PGPL baseline. Hence, we use the results reported in the original research paper to avoid re-implementation bias. However, their setup is slightly different than the setup of our experiments as they, e.g., use back-translation for data augmentation.  

\paragraph*{Performance of \ourmodel versus Distil-\ourmodel} 
To assess the usefulness of \ourmodel for limited environments we developed Distil-\ourmodel and evaluated its performance. Overall, Distil-\ourmodel performs well in comparison to \ourmodel even though its student model is considerably smaller. In particular, the performance gap between Distil-\ourmodel and \ourmodel on \ag and \yahoo is less than 0.4\% in most cases.

For a fair comparison between \ourmodel and Distil-\ourmodel on \bonnier, one should only compare Distil-\ourmodel(SE-multi) with \ourmodel(SE-multi), using a multilingual language model for the student in both cases. However, among all of the available options the unilingual \ourmodel(SE-SE) is clearly the most reasonable one considering its size and performance.

\paragraph{The effect of objective masking on classification performance}

The results in Tables~\ref{tab:all_results} and~\ref{tab:all_results_bonnier} indicate that pre-trained BERT with objective masking consistently outperforms the BERT classifier without pre-training. In particular, in the 10-shot case, pre-trained BERT shows a significant improvement of 2.7\% for \yahoo, 2.6\% for \ag, and 0.5\% for \bonnier in terms of accuracy. Additionally, when compared to pre-trained BERT with random masking, the objective masking approach yields a performance improvement of 0.2\% to 1.3\% across all datasets. This observation highlights that the superiority of pre-trained BERT with objective masking is not solely attributed to domain adaptation but also to the effectiveness of the topic-based masking approach. Looking at Table~\ref{tab:masking-exmpls}, which shows some examples of how the two masking strategies choose words for masking, this seems intuitively reasonable: the words chosen according to the LDA model are obviously semantically more important.

Next, we compare \ourmodel and \ourmodelM. The performance of \ourmodelM is superior to that of \ourmodel on all datasets in all cases. Notably, in the 10-shot case, \ourmodelM exhibits an absolute increase in accuracy of 1.7\% for \yahoo, 0.4\% for \ag, and 0.5\% for \bonnier when compared to \ourmodel. \ourmodelM with objective masking also outperforms \ourmodelM with random masking. In most cases the advantage is significant, the exception being \bonnier, on which the difference is small. 
Similar findings were observed for pre-trained DistilBERT with objective masking and Distil-\ourmodelM, where objective masking gave improved results. However, the impact of objective masking varies depending on the dataset and the complexity of the classification task.

Tables~\ref{tab:all_results} and~\ref{tab:all_results_bonnier} show that objective masking is particularly advantageous when there is a scarcity of supervised information. We trained BERT/DistilBERT classifiers using labeled data only, in contrast to \ourmodel, which incorporates both labeled and pseudo-labeled data. As a result, we observe more substantial improvements with objective masking for the BERT/DistilBERT classifiers compared to their \ourmodel/Distil-\ourmodel counterparts. In three distinct scenarios, the most significant enhancement is observed for the 10-labeled case in both BERT and \ourmodel. Additionally, as the amount of labeled data increases, the distinction between objective masking and random masking becomes less pronounced for both BERT and \ourmodel. We conjecture that this can be attributed to the increasing influence of fine-tuning during model training. 

\begin{table*}
\renewcommand{\tabularxcolumn}[1]{m{#1}}
\resizebox{\textwidth}{!}{%
\begin{tabularx}{1.2\linewidth}{lXp{.1\linewidth}p{.4\linewidth}}
\toprule
\textbf{Dataset} &
  \textbf{Sentence (tokenized)} &
  \textbf{\begin{tabular}{@{}c@{}}Masking \\ policy\end{tabular}} &
  \textbf{Selected words} \\\midrule
\yahoo &
  '{[}CLS{]}', 'do', 'u', 'think', 'that', 'golf', 'is', 'the', 'most', 'boring', 'high', 'paid', 'sport', 'to', 'watch', 'on', 'tv', '?', 'did', 'you', 'ever', 'watch', 'curling', '?', '{[}SEP{]}' &
  \multicolumn{2}{l}{\begin{tabular}{@{}m{.1\linewidth}m{.4\linewidth}@{}}
  threshold &
  'tv', 'high', 'boring', 'watch' \\\midrule
  relevance &
  'tv', 'high', 'boring', 'watch' \\\midrule
  random &
  'on', 'you', 'that', 'paid'\end{tabular}}
  \\\midrule
\yahoo &
  '{[}CLS{]}', 'what', 'does', 'e', '=', 'mc', '\#\#2', 'mean', '?', 'it', 'is', 'an', 'equation', 'by', 'albert', 'einstein', 'showing', 'that', 'energy', 'and', 'mass', 'are', 'interchange', '\#\#able', '.', 'e', 'is', 'energy', 'm', 'is', 'mass', 'c', 'is', 'the', 'speed', 'of', 'light', 'thus', 'energy', 'equals', 'the', 'amount', 'of', 'mass', 'multiplied', 'by', 'the', 'speed', 'of', 'light', 'squared', '.', '{[}SEP{]}' &
  \multicolumn{2}{l}{\begin{tabular}{@{}m{.1\linewidth}m{.4\linewidth}@{}}
  threshold &
  'einstein', 'squared', 'speed', 'mass',  'speed', 'albert', 'mass', 'multiplied' \\\midrule
  relevance &
  'einstein', 'squared', 'speed', 'mass', 'speed', 'albert', 'mass', 'multiplied' \\\midrule
  random &
  'the', 'of', 'amount', 'c', 'e', 'is', 'is', 'does'\end{tabular}}
  \\\midrule
\ag &
  '{[}CLS{]}', 'wages', 'rose', 'faster', 'than', 'expected', 'in', 'the', 'june', '-', 'august', 'period', 'but', 'analysts', 'say', 'the', 'increases', 'are', 'still', 'not', 'high', 'enough', 'to', 'cause', 'inflation', 'worries', 'at', 'the', 'bank', 'of', 'england', '.', '{[}SEP{]}' &
  \multicolumn{2}{l}{\begin{tabular}{@{}p{.1\linewidth}p{.4\linewidth}@{}}
  threshold &
  'bank', 'increases', 'inflation', 'august', 'england' \\\midrule
  relevance &
  'bank', 'increases', 'inflation', 'england', 'worries' \\\midrule
  random &
  'wages', 'high', 'analysts', 'than', 'bank'\end{tabular}}
  \\\midrule
\ag &
  '{[}CLS{]}', 'italian', 'anti', '-', 'mafia', 'magistrates', 'ordered', 'the', 'arrest', 'of', '65', 'people',  'as', 'part', 'of', 'a', 'massive', 'police', 'sw', '\#\#oop', 'in', 'naples', 'early', 'today', 'in', 'a', 'bid', 'to', 'staunch', 'the', 'blood', '\#\#lett', '\#\#ing', 'in', 'a', 'turf', 'war', 'which', 'has', 'killed', 'more', 'than', '120', 'people', ',', 'interior', 'minister', 'giuseppe', 'pisa', '\#\#nu', 'said', '.', '{[}SEP{]}' &
  \multicolumn{2}{l}{\begin{tabular}{@{}p{.1\linewidth}p{.4\linewidth}@{}}
  threshold &
  'people', 'bid', 'part', 'magistrates', 'mafia', 'massive', 'police', 'war' \\\midrule
  relevance &
  'people', 'bid', 'part', 'massive', 'police', 'war', 'said', 'italian' \\\midrule
  random &
  'said', 'which', '65', 'a', '120', 'today', 'naples', 'people'\end{tabular}}
  \\\bottomrule
\end{tabularx}%
}
\caption{Examples of how different masking policies choose words for masking in pre-training\label{tab:masking-exmpls}}
\end{table*}

The optimal choice of the relevance parameter $\lambda$ (see Section \ref{sec:choose list}) for making topic word lists depends on the characteristics of the dataset. In complex datasets with numerous topics and overlapping word distributions between many of them, a smaller $\lambda$ value proves more beneficial. On the other hand, for simpler classification tasks, such as those with fewer classes to be recognized and well-separated classes without overlaps, a larger $\lambda$ performs better. As evidenced by the results in Tables \ref{tab:all_results} and \ref{tab:all_results_bonnier} for \yahoo and \bonnier, lambda values of 0.2 and 0.1 are more effective than 0.7, while for \ag, a lambda value of 0.7 yields better results. The visualization of the topic models for these datasets via pyLDAvis\footnote{\url{https://pypi.org/project/pyLDAvis/}} shows that \yahoo and \bonnier both contain several overlapping topics meaning there are a lot of common words between these topics. Thus, these datasets favor a low $\lambda$ value which aligns with our previous explanation. In the case of BERT, smaller values of $\lambda$ consistently prove to be superior. We conjecture that with limited labeled data, a smaller $\lambda$ helps the model distinguish between different classes by learning the topic-specific words.

Moreover, the quality of data used for pre-training significantly influences the effectiveness of objective masking. For instance, objective masking demonstrates greater efficacy on \yahoo compared to \ag. Specifically, for \yahoo, the accuracy of \ourmodelM improved by 1.7\% compared to the accuracy of \ourmodel in the case of 10 labeled examples per class, whereas for \ag, the improvement was 0.4\%. Based on some additional experiments that we performed, we conclude that this variation can most likely be attributed to the fact that \ag consists of short texts and is smaller overall, comprising only 3.6M words in total, whereas \yahoo consists of longer texts with 12.2M words in total. Consequently, \ag offers considerably less context for BERT during pre-training than \yahoo does. We tested this hypothesis by running experiments on the text of \ag documents only, stripping away the titles. As expected, the performance dropped. Table \ref{tab:ag_pretrain} presents the results of these experiments.

\begin{table}[ht!]
\newcommand{\two}[1]{\textbf{\begin{tabular}{@{}l@{}}#1\end{tabular}}}
\centering
\begin{tabular}{@{}lccc@{}}
\toprule
\two{Pre-\\training\\data} & \two{\ourmodelM\\(random)} & \two{\ourmodelM\\(relev.-0.2)} & \two{\ourmodelM\\(relev.-0.7)} \\\midrule
Text                       & 87.5                               & 87.8                                      & 87.9                                      \\
Text+title                 & 88.0                               & 88.4                                      & 88.5  \\\bottomrule                                   
\end{tabular}
\caption{Comparison of the accuracy of different \ourmodelM versions with encoders pre-trained under two different settings on \ag for the 10-labeled case. We either use only the text or the concatenation of the text and the document title. The average accuracy of five runs with distinct random seeds is reported.\label{tab:ag_pretrain}}
\end{table}

As it can be seen in Table~\ref{tab:bonnier_stat}, \bonnier exhibits a significant class imbalance, with a majority of its documents belonging to the Sports category. This imbalance poses a considerable challenge for LDA to generate topics that accurately align with the actual categories. Table~\ref{tab:bonnier_topics} presents the topics obtained from LDA for \bonnier. 
As depicted in the table, certain topics lack coherence and cannot be accurately matched with a specific class in the dataset. Additionally, more than one topic is associated with the `Sport' category, while no matches were found for the topics `Personligt', `Vetenskap \& teknologi', `Religion \& tro', `Miljö', `Konflikter, krig \& terrorism', and `Arbetsmarknad'. As a consequence, the topic word lists generated by LDA do not adequately capture all the diverse topic words in the dataset, resulting in limited effectiveness of objective masking compared to random masking. Moreover, as noted in the Swedish BERT paper~\citep{Swedish-bert}, the training data for Swedish BERT heavily leans towards newspaper text, leading to a similarity between the data distribution of Swedish BERT training data and \bonnier. This similarity, in turn, restricts the effectiveness of pre-training.

\begin{sidewaystable}
\resizebox{\textwidth}{!}{%
\begin{tabular}{@{}ll@{}}
\toprule
  \textbf{Topic words} &
  \textbf{Best match} \\\midrule
  kund, butik, företag, bolag, konkurs, företagsnamn, produkt, köpare, fabrik, säte&
  Ekonomi, näringsliv \& finans \\\midrule
  trafikverk, trafik, hastighet, cyklist, cykelväg, järna, tågtrafik, resenär, järnväg, cykelbana&
  Samhälle \& välfärd\\\midrule
  patient, minska, regering, region, parti, procent, politik, vård, politisk, införa&
  Politik\\\midrule
  vatten, utställning, snö, sjö, sol, vädr, träd, regn, väder, vind&
  Väder\\\midrule
  säsong, klubb, spelare, träning, trupp, träna, kontrakt, sportchef, stars, nyförvärv&
  Sport\\\midrule
  faktafel, hofors, flygplats, orientering, arbetsplatsolycka, ockelbo, flygplatse, anmäl, euro, lott, auktion, aktieäg&
  - \\\midrule
  bollnäs, söderhamn, edsby, carlström, hudik, aida, edsbyn, vänersborg, falbygd, johannesson&
  - \\\midrule
  vaccin, ericsson, vaccinera, dos, habo, vaccination, vaccinering, indycar, rosenqvist, wolley&
  Hälsa \& sjukvård\\\midrule
  elev, förskola, skola, lokalerna, rektor, bygglov, undervisning, högskola, lärare, grundskola &
  Skola \& utbildning\\\midrule
  herr, bortalag, tabell, ibk, leksand, brynäs, köping, innebandy, målgörare, ibf&
  Sport\\\midrule
  polis, räddningstjänst, larm, larma, brande, brand, ambulans, olycka, presstalesperson, förar&
  Olyckor \& katastrofer\\\midrule
  kvinna, åtala, tingsrätt, döma, fängelse, åklagar, sexuell, brott, våld, våldtäkt&
  Brott \& straff\\\midrule
  liv, familj, bök, mamma, pappa, läsare, äta, son, språk, roman&
  Livsstil \& fritid\\\midrule
  tävling, lopp, final, tävla, set, tävlingarna, medalj, brons, åkare, sprint &
  Sport\\\midrule
  domare, gif, vsk, boll, degerfors, hörna, brage, kjäll, skutskär, kvarnsved&
  Sport\\\midrule
  musik, scen, band, föreställning, konsert, låt, artist, kyrka, sång, festival&
  Kultur \& nöje\\\midrule
  varg, hus, katt, arkitektur, kvadratmet, bostadsrätt, revir, hyresgäst, skyddsjakt, hyresrätt&
  -\\\bottomrule
\end{tabular}%
}
\caption{Top 10 words (based on relevance score with $\lambda=0.1$) for the topics identified by LDA in \bonnier and the best matches we found for them from the real classes in the dataset. Some topics lack sufficient coherence to be matched to specific classes in the dataset.\label{tab:bonnier_topics}}
\end{sidewaystable}

\subsection{The Impact of Objective Masking on Domain-Specific Tasks (\question5)\label{sec:key_factors}}

When the dataset deviates significantly from the BERT training data and incorporates domain-specific information, such as medical documents, the distinction between the effect of objective masking and random masking on the classification performance is more pronounced than if the classification task relies mostly on general language understanding. In the latter case, BERT's pre-trained knowledge from a large general corpus may be sufficient to handle the task.

To gain deeper insights into this aspect, we compare the impact of objective masking and random masking on the classification performance of the BERT and \ourmodelM classifiers on the Medical Abstracts dataset\footnote{\url{https://www.kaggle.com/datasets/chaitanyakck/medical-text}}~\citep{medic-dataset-paper}. Table~\ref{tab:medic_stat} shows the class distribution within this dataset.

\begin{table}[ht!]
\centering
\begin{tabularx}{\linewidth}{@{}X@{}c@{\ \ }c@{\ \ }c@{}}
\toprule
\textbf{Class name}                      & \textbf{Training} & \textbf{Test} & \textbf{Total} \\\midrule
Neoplasms                       & 2530       & 633    & 3163  \\
Digestive system diseases       & 1195       & 299    & 1494  \\
Nervous system diseases         & 1540       & 385    & 1925  \\
Cardiovascular diseases         & 2441       & 610    & 3051  \\
General pathological conditions & 3844       & 961    & 4805  \\\midrule
Total                          & 11550      & 2888   & 14438\\\bottomrule
\end{tabularx}%
\caption{Class distribution within Medical Abstracts\label{tab:medic_stat}}
\end{table}

We create an LDA topic model for the dataset with 5 topics and make two relevance-based topic word lists: one with $\lambda=0.2$, comprising 500 words for each of the 5 topics, and another with $\lambda=0.7$ consisting of 700 words per topic. For the experiments, we allocated 0.1 of the training data as the validation set. In three distinct settings, we divided the remaining data into labeled and unlabeled datasets, using proportions of 0.01, 0.1, and 0.3 for the labeled dataset, while the rest was assigned to the unlabeled dataset. Additionally, we applied the same augmentation method used for other datasets to generate an augmented version of the dataset. The results of these experiments are displayed in Table~\ref{tab:medic}.

\begin{table}[ht!]
\centering
\begin{tabular}{@{}lccc@{}}
\toprule
Model                    & 0.01       & 0.1        & 0.3        \\
\midrule
BERT                     & 49.2 & 59.0 & 61.0          \\
BERT (random)            & 51.0 & 60.5 & 61.9          \\
BERT (relevance-0.2)     & 54.7 & 61.4 & 62.1 \\
BERT (relevance-0.7)     & 53.8 & 61.8 & 62.2 \\
\midrule
\ourmodel                  & 53.7          & 60.8         & 62.8         \\
\ourmodelM (random)         & 55.2          & 62.2         & 63.8          \\
\ourmodelM (relevance-0.2) & 56.4          & \textbf{62.6}          & \textbf{64.2}          \\
\ourmodelM (relevance-0.7) & \textbf{57.1}          & \textbf{62.6}          & \textbf{64.2}          \\
\bottomrule
\end{tabular}%
\caption{Comparison of test accuracy for different variations of BERT and \ourmodel
pretrained with new domain-specific topic word lists on Medical Abstracts dataset. The average accuracy of five runs with distinct random seeds is reported.\label{tab:medic}}
\end{table}

Comparing the results in Table~\ref{tab:medic} with those in Table \ref{tab:all_results}, we observe that objective masking outperforms random masking by a larger margin for the Medical Abstracts dataset in comparison to the other datasets, particularly in the BERT setting. Specifically, in the case with the minimal labeled data, the BERT (relevance) and \ourmodelM(relevance) models demonstrate superior performance over BERT (random) and \ourmodelM(random), achieving 3.7\% and 1.9\% increase in accuracy respectively. In contrast, the corresponding accuracy improvements on \yahoo and \ag are (1.3\%, 1.2\%) and (0.7\%, 0.5\%), respectively.

We note also that the topics in Medical Abstracts are fairly well separated and $\lambda=0.7$ is preferable in all cases except the 0.01 labeled case of the BERT classifier. In this case, the supervised data is extremely limited, so using more specific words for masking could potentially enable the model to concentrate on the distinct characteristics of each topic and enhance its ability to classify instances belonging to those specific topics.

\subsection{Comparison of LDA and TF-IDF for Topic Word Extraction (\question6)}
To compare the LDA-based methods for generating topic word lists with simpler techniques that do not rely on topic models, we present alternative versions of our models that use TF-IDF to identify topic words within the corpus. Specifically, we sort the words based on their average TF-IDF scores across all documents and select the top words. The preprocessing step remains consistent with the LDA-based method: we remove general stop words and a few of the most frequent words in the dataset, keeping only nouns, verbs, and adjectives. After preprocessing, we calculate the average TF-IDF score for all words in the vocabulary and choose the top $N$ words. The value of $N$ matches the length of the relevance-based list that has been proven to be more effective for each dataset (e.g., the list with $\lambda=0.2$ for \yahoo, the list with $\lambda=0.7$ for \ag, and the list with $\lambda=0.1$ for \bonnier). Tables~\ref{tab:tfidf} and \ref{tab:tf-idf_bonnier} show the results of these comparisons.

\begin{table*}
\resizebox{\textwidth}{!}{
\begin{tabular}{@{}l@{\hspace{1.5ex}}lllll@{\hspace{1.5ex}}llll@{}}
\toprule
 &
   &
  \multicolumn{3}{@{}c@{}}{\begin{tabular}{@{}c@{}}Labeled examples\\ per class\end{tabular}} &
   &
   &
  \multicolumn{3}{@{}c@{}}{\begin{tabular}{@{}c@{}}Labeled examples\\ per class\end{tabular}} \\\cmidrule(lr){3-5}\cmidrule(lr{0pt}){8-10}
 &
  Model &
  10 &
  200 &
  2500 &
&
  Model &
  10 &
  200 &
  2500 \\\midrule
\multirow{6}{*}{\rotatebox{90}{\yahoo}} &
  BERT (relevance-0.2) &
  62.9 &
  71.5 &
  74.4 &
 \multirow{6}{*}{\rotatebox{90}{\ag}} &
  BERT (relevance-0.2) &
  84.5 &
  89.2 &
  91.9 \\
&
  BERT (relevance-0.7) &
  62.7 &
  71.5 &
  74.3 &
&
  BERT (relevance-0.7) &
  84.3 &
  89.1 &
  91.9 \\
&
  BERT (tf-idf) &
  62.5 &
  71.5 &
  74.2 &
& BERT (tf-idf) &
  84.3 &
  89.1 &
  91.7  \\\cmidrule(lr){2-5}\cmidrule(lr){7-10}
&  \ourmodelM (relevance-0.2) &
   \textbf{66.3}  &
   \textbf{72.9}  &
   \textbf{75.1}  &
&
  \ourmodelM (relevance-0.2) &
   88.4  &
   90.1  &
   \textbf{92.2}  \\
&
  \ourmodelM (relevance-0.7) &
   66.1  &
   72.8  &
   75.0  &
   &
  \ourmodelM (relevance-0.7) &
   \textbf{88.5} &
   \textbf{90.2}  &
   \textbf{92.2} \\
&
  \ourmodelM (tf-idf) &
   65.4  &
   72.8  &
   75.0  &
&
  \ourmodelM (tf-idf) &
   88.4  &
   \textbf{90.2}  &
   92.1 \\
\bottomrule
\end{tabular}
}
\caption{Comparison of LDA and TF-IDF for \yahoo and \ag\label{tab:tfidf}}
\end{table*}

\begin{table*}
\centering
\begin{tabular}{llccc}
\toprule
& 
& 
\multicolumn{3}{@{}c@{}}{\begin{tabular}{@{}c@{}}Labeled examples\\ per class\end{tabular}} \\\cmidrule(lr{0pt}){3-5}
Dataset &   Model   &   0.01    &   0.1    &   0.3\\\midrule
\multirow{6}{*}{\bonnier} & BERT-SE  (relevance-0.1)   &  82.2     &  85.8     &  87.2  \\
& BERT-SE  (relevance-0.7)   &  82.0     &  85.7     &  87.1  \\
& BERT-SE  (tf-idf)   &  82.0     &  85.6     &  87.0  \\\cmidrule(lr{0pt}){2-5}
& \ourmodelM (SE-SE) (relevance-0.1) &    \textbf{83.8}     &    \textbf{86.7}    &   \textbf{87.7}    \\
& \ourmodelM (SE-SE) (relevance-0.7)    &    83.6     &     86.6   &   87.6 \\
& \ourmodelM (SE-SE) (tf-idf)    &    83.5     &     86.6   &   87.6 \\       
\bottomrule 
\end{tabular}
\caption{Comparison of LDA and TF-IDF for \bonnier\label{tab:tf-idf_bonnier}}
\end{table*}

As can be seen in Tables \ref{tab:tfidf} and \ref{tab:tf-idf_bonnier}, masking with topic words is slightly less effective if the selection is based on TF-IDF instead of LDA. As could be expected given the relatively good performance of even random masking, the difference is small if there is a sufficient amount of labeled data, but if the labeled data is severely limited the effect is more pronounced. We hypothesize that this superiority can be attributed to the fact that the topic model considers the underlying structure of the dataset, whereas TF-IDF relies on individual documents. Nevertheless, the TF-IDF approach does identify a certain number of true topic words, making it a reasonable compromise when facing resource constraints such as time and computational power. However, it should also be noted that the LDA-based method offers flexibility in choosing between highly topic-specific words and more general ones, catering to the specific needs of the analysis, while the TF-IDF method offers less control over the generated lists.

\subsection{Effect of Objective Masking on Reliability and Interpretability (\question7)} 

\citet{Moon2020MASKERMK} illustrated that fine-tuning a text classifier using masked keyword regularization helps it consider context rather than solely relying on certain keywords, which leads to improved out-of-distribution detection and cross-domain generalization. Inspired by this research, we conducted a qualitative study to examine how our proposed masking strategy impacts the reliability of \ourmodelM. To do so, we performed case studies on \yahoo and \ag, using \ourmodelM~(relevance).  
We selected samples from the test set of the datasets that were correctly classified by \ourmodelM(relevance) and arranged them in order of their predicted probability of belonging to the correct class. Afterwards, we chose 20 samples classified with lower probabilities and examined cases that were misclassified by both \ourmodel and \ourmodelM~(random) to identify the factors that contribute to the accurate classification by \ourmodelM~(relevance). Tables~\ref{tab:att_yahoo} and~\ref{tab:att_ag} show some of these examples.

 In the upper example of Table~\ref{tab:att_yahoo}, ``gun'' was the most commonly attended word across all models. However, \ourmodelM (relevance) correctly predicted the class ``Sports'' by considering the context and specifically the word ``safe''. In contrast, \ourmodelM (random) predicted ``Business'' by taking into account the words ``gun'' and ``steel''. Furthermore, the prediction of the class ``Politics \& Government'' by \ourmodel indicates a disregard for the context. This trend is consistent across models in the second example as well, where \ourmodelM (relevance) accurately associates ``digital technology'' with the ``Computers \& Internet'' category. Moreover, Table~\ref{tab:att_ag} shows examples of \ag where \ourmodelM (relevance) predicts the correct topic for the document by considering not only the keywords but also their contextual information. These observations indicate that pre-training with objective masking on domain-specific datasets teaches the model the contexts in which a given keyword may appear, thus enabling it to account for contextual factors during classification.

It is also worth noting that both \ourmodelM variations exhibit higher interpretability due to their attention to more relevant and informative words, as compared to \ourmodel.
\begin{table*}
\small
\resizebox{\textwidth}{!}{%
\begin{tabular}{p{1.5cm}p{2cm}p{10.5cm}p{1.7cm}}
\toprule
Class & Model & Text & Prediction\\
\midrule
\multirow{9}{*}{Sports} & \ourmodel & \colorbox{teal!12.194749218982249}{where} \colorbox{teal!9.860234126655268}{can} \colorbox{teal!27.694741235836567}{i} \colorbox{teal!20.98226905329776}{find} \colorbox{teal!51.53296925120374}{a} \colorbox{teal!39.2273949627111}{che} \colorbox{teal!20.745614401898923}{\#\#ep} \colorbox{teal!100.0}{gun} \colorbox{teal!61.71302339974043}{safe} \colorbox{teal!37.56131245956156}{made} \colorbox{teal!26.27227012532018}{of} \colorbox{teal!71.05147152183847}{steel} ? \colorbox{teal!10.675117523308884}{i} \colorbox{teal!1.2178013124291496}{don} \colorbox{teal!1.7094309841850082}{t} \colorbox{teal!0.0}{think} \colorbox{teal!47.69821580495579}{you} \colorbox{teal!7.8457500393455035}{should} \colorbox{teal!24.17637505882818}{use} \colorbox{teal!26.452647401181963}{the} \colorbox{teal!28.748440435469192}{words} \colorbox{teal!48.77233665153417}{cheap} \colorbox{teal!26.046822289855093}{and} \colorbox{teal!38.879462863546756}{safe} \colorbox{teal!7.630754422472632}{in} \colorbox{teal!16.88338562927336}{the} \colorbox{teal!21.833842607340998}{same} \colorbox{teal!37.91770289164742}{sentence} \colorbox{teal!40.66885835694698}{it} \colorbox{teal!18.674121150698223}{sounds} \colorbox{teal!40.81414780680179}{dangerous}  & Politics \& Government \\\cmidrule(lr){2-4}
                       & \ourmodelM (random) & \colorbox{teal!23.606716663213096}{where} \colorbox{teal!12.540950445364507}{can} \colorbox{teal!12.090144282370876}{i} \colorbox{teal!29.612370241336844}{find} \colorbox{teal!44.70211399812396}{a} \colorbox{teal!27.743245025667857}{che} \colorbox{teal!9.504769277528538}{\#\#ep} \colorbox{teal!100.0}{gun} \colorbox{teal!54.1964783258627}{safe} \colorbox{teal!29.84270112103368}{made} \colorbox{teal!26.193354197913415}{of} \colorbox{teal!95.15766065493125}{steel} ? \colorbox{teal!2.927302165584919}{i} \colorbox{teal!0.0}{don} \colorbox{teal!5.299659113661883}{t} \colorbox{teal!7.900714171420385}{think} \colorbox{teal!35.85337423828139}{you} \colorbox{teal!21.211548989691607}{should} \colorbox{teal!30.983814050496015}{use} \colorbox{teal!25.898324123764144}{the} \colorbox{teal!37.0088948614172}{words} \colorbox{teal!38.57919367613211}{cheap} \colorbox{teal!36.42237329161335}{and} \colorbox{teal!39.574857923616094}{safe} \colorbox{teal!13.175312722693585}{in} \colorbox{teal!10.443126027860124}{the} \colorbox{teal!32.12099085745993}{same} \colorbox{teal!52.20546000258993}{sentence} \colorbox{teal!47.446867015229024}{it} \colorbox{teal!36.06793441629198}{sounds} \colorbox{teal!56.53234641359772}{dangerous} & Business \& Finance\\\cmidrule(lr){2-4}
                       & \ourmodelM(relevance) & \colorbox{teal!26.27666652884865}{where} \colorbox{teal!14.6014400622062}{can} \colorbox{teal!26.796548019552173}{i} \colorbox{teal!26.525883494802223}{find} \colorbox{teal!66.28332556406458}{a} \colorbox{teal!19.834502450055314}{che} \colorbox{teal!4.528575737179982}{\#\#ep} \colorbox{teal!100.0}{gun} \colorbox{teal!60.900218988684315}{safe} \colorbox{teal!47.49314651125527}{made} \colorbox{teal!27.056896895504313}{of} \colorbox{teal!56.17942556860134}{steel} ? \colorbox{teal!13.128579534569045}{i} \colorbox{teal!1.3474254323212107}{don} \colorbox{teal!1.3738969900841418}{t} \colorbox{teal!0.0}{think} \colorbox{teal!33.12236530255815}{you} \colorbox{teal!11.04763694452612}{should} \colorbox{teal!22.18633728989333}{use} \colorbox{teal!18.142573386466022}{the} \colorbox{teal!29.287380903498228}{words} \colorbox{teal!43.617970120463}{cheap} \colorbox{teal!28.758975989676784}{and} \colorbox{teal!33.674214066156075}{safe} \colorbox{teal!10.96462229316991}{in} \colorbox{teal!11.601829501608064}{the} \colorbox{teal!11.921432730051242}{same} \colorbox{teal!39.92362811778507}{sentence} \colorbox{teal!57.040297218447996}{it} \colorbox{teal!32.993289563988164}{sounds} \colorbox{teal!58.455901163308255}{dangerous}  & Sports\\\midrule
\multirow{9}{*}{\begin{tabular}{@{}c@{}}Computers \\ \& Internet\end{tabular}} & \ourmodel & \colorbox{purple!54.162687242281926}{what} \colorbox{purple!100.0}{are} \colorbox{purple!75.34638088664418}{the} \colorbox{purple!91.02035883232776}{differences} \colorbox{purple!37.08093167000251}{between} \colorbox{purple!51.046571130693074}{digital} \colorbox{purple!75.08608879693975}{technology} \colorbox{purple!65.03101539798159}{and} \colorbox{purple!0.0}{information} \colorbox{purple!72.99958897520669}{technology} & Business \& Finance\\\cmidrule(lr){2-4}
                    & \ourmodelM (random) & \colorbox{purple!0.0}{what} \colorbox{purple!58.3594627470818}{are} \colorbox{purple!31.942640375110592}{the} \colorbox{purple!100.0}{differences} \colorbox{purple!9.43346972576994}{between} \colorbox{purple!99.23276879541389}{digital} \colorbox{purple!84.7199939941658}{technology} \colorbox{purple!70.58361520060133}{and} \colorbox{purple!38.7488181985027}{information} \colorbox{purple!89.99426069369882}{technology} & Science \& Mathematics\\\cmidrule(lr){2-4}
                    & \ourmodelM (relevance) & \colorbox{purple!0.0}{what} \colorbox{purple!13.483209069305492}{are} \colorbox{purple!9.92504489847101}{the} \colorbox{purple!70.3354764497291}{differences} \colorbox{purple!14.075717074795346}{between} \colorbox{purple!100.0}{digital} \colorbox{purple!78.40634879960776}{technology} \colorbox{purple!47.30721740307084}{and} \colorbox{purple!22.738702625362627}{information} \colorbox{purple!52.53008071831161}{technology} & Computers \& Internet\\\bottomrule
\end{tabular}%
}
\caption{Visualization of selected samples from \yahoo. The color saturation indicates the average attention to the word from other words in the sentence.\label{tab:att_yahoo}}
\end{table*}

\begin{table*}
\small
\resizebox{\textwidth}{!}{%
\begin{tabular}{p{1.5cm}p{2cm}p{10.5cm}p{1.7cm}}
\toprule
Class & Model & Text & Prediction\\
\midrule
\multirow{3}{*}{Business} & \ourmodel & \colorbox{teal!19.83854102450426}{how} \colorbox{teal!48.013431304908984}{will} \colorbox{teal!8.22194182041401}{companies} \colorbox{teal!33.127558047109254}{and} \colorbox{teal!33.33585869847507}{investors} \colorbox{teal!52.14366249649134}{fare} \colorbox{teal!30.011915036503247}{if} \colorbox{teal!29.081786678349104}{the} \colorbox{teal!100.0}{storm} \colorbox{teal!16.165467380554514}{spawn} \colorbox{teal!0.0}{\#\#s} \colorbox{teal!1.9816473810267077}{moderate} \colorbox{teal!48.05128030241422}{damage}  & Sci/Tech \\\cmidrule(lr){2-4}
                       & \ourmodelM (random) & \colorbox{teal!29.506592019120426}{how} \colorbox{teal!65.47426718141377}{will} \colorbox{teal!70.50797813850943}{companies} \colorbox{teal!70.89923231100046}{and} \colorbox{teal!99.71061966606781}{investors} \colorbox{teal!47.32532313428814}{fare} \colorbox{teal!56.27664752106259}{if} \colorbox{teal!100.0}{the} \colorbox{teal!88.91955636166455}{storm} \colorbox{teal!0.0}{spawn} \colorbox{teal!11.782767255999087}{\#\#s} \colorbox{teal!7.608754231489902}{moderate} \colorbox{teal!48.08260976441932}{damage}  & Sci/Tech \\\cmidrule(lr){2-4}
                       & \ourmodelM (relevance) & \colorbox{teal!30.737263201298614}{how} \colorbox{teal!51.83205471659916}{will} \colorbox{teal!49.68088607704985}{companies} \colorbox{teal!66.51797080605638}{and} \colorbox{teal!67.48417409059239}{investors} \colorbox{teal!16.56608504819497}{fare} \colorbox{teal!53.77289431881833}{if} \colorbox{teal!100.0}{the} \colorbox{teal!89.29903982652047}{storm} \colorbox{teal!1.9269170246669325}{spawn} \colorbox{teal!23.115024584961876}{\#\#s} \colorbox{teal!0.0}{moderate} \colorbox{teal!47.517098937239105}{damage} & Business\\\midrule
\multirow{3}{*}{Business} & \ourmodel & \colorbox{purple!10.143625956870055}{in} \colorbox{purple!31.615742302087774}{recent} \colorbox{purple!0.0}{years} \colorbox{purple!62.757821995125774}{hundreds} \colorbox{purple!47.28975902464477}{of} \colorbox{purple!13.018459717600045}{multinational} \colorbox{purple!27.1214994304913}{companies} \colorbox{purple!64.89316144139788}{have} \colorbox{purple!46.999126647280036}{set} \colorbox{purple!28.1828456981413}{up} \colorbox{purple!71.2252709143319}{research} \colorbox{purple!100.0}{laboratories} \colorbox{purple!67.49758197425439}{in} \colorbox{purple!32.03155523598022}{china} & Sci/Tech\\\cmidrule(lr){2-4}
                    & \ourmodelM (random) & \colorbox{purple!13.432399516078508}{in} \colorbox{purple!8.967782674602404}{recent} \colorbox{purple!0.0}{years} \colorbox{purple!43.12133413266794}{hundreds} \colorbox{purple!67.77470826087114}{of} \colorbox{purple!25.8358957648468}{multinational} \colorbox{purple!47.302883305548654}{companies} \colorbox{purple!29.44120539898919}{have} \colorbox{purple!60.16436917976987}{set} \colorbox{purple!49.4148714643546}{up} \colorbox{purple!79.1182235088389}{research} \colorbox{purple!100.0}{laboratories} \colorbox{purple!65.1804113812766}{in} \colorbox{purple!50.73430615715686}{china} & Sci/Tech\\\cmidrule(lr){2-4}
                    & \ourmodelM (relevance) & \colorbox{purple!8.683017320683605}{in} \colorbox{purple!19.427601091621387}{recent} \colorbox{purple!0.0}{years} \colorbox{purple!44.30915205705593}{hundreds} \colorbox{purple!74.98803599436793}{of} \colorbox{purple!30.72908358329422}{multinational} \colorbox{purple!55.65637746814286}{companies} \colorbox{purple!31.197708381934703}{have} \colorbox{purple!55.793137501694325}{set} \colorbox{purple!39.50634530287145}{up} \colorbox{purple!90.61193622024713}{research} \colorbox{purple!100.0}{laboratories} \colorbox{purple!78.07585594507346}{in} \colorbox{purple!47.73292466687506}{china} & Business\\\bottomrule
\end{tabular}%
}
\caption{Visualization of selected samples from the \ag. The color saturation indicates the average attention to the word from other words in the sentence.\label{tab:att_ag}}
\end{table*}

\subsection{Effect of Batch-size and Number of GPUs on the Performance (\question8)\label{sec:gpu}}

Since the performance of a model depends on the training data, it varies when the amount of training samples used to train the model is dynamic. In case of a teacher-student model, the dynamic change of the number of training examples affects the information sharing between teacher and student (including the student feedback for meta pseudo-labeling approaches). Therefore, the dependency of the performance on the number of training examples may be suspected to be even stronger in such a model.
To study this, we experiment with the batch size and the number of GPUs being used in two main scenarios, namely 10 labeled samples and 200 labeled samples for the labeled set of \ag. We base our analysis on \ag for two reasons. First, it is a publicly available dataset, meaning that the results of the analysis can be replicated by others. Second, Section \ref{sec:result_analysis} concluded that \ourmodelM performs better on long-text datasets. As seen in Table \ref{tab:data_set}, \ag consists of shorter documents than the other two datasets. Hence, any differences revealed by comparing the average performance of \ourmodel and \ourmodelM on \ag can be expected to carry over to other cases.

Tables~\ref{tab:batch_10}--\ref{tab:gpu_200} show the results of these experiments. In Tables~\ref{tab:batch_10} and~\ref{tab:batch_200}, the \emph{global} batch size
is the product of the local batch size and the number of GPUs used for the experiment. The reported accuracies are averages over multiple experiments with varying numbers of GPUs and local batch sizes, keeping the global batch size constant as shown in the Table. Tables~\ref{tab:batch_10} and~\ref{tab:batch_200} show a rather distinctive behavior: in the 10-sample experiments, the smallest batch sizes favor \ourmodel, but at a certain point \ourmodelM starts to perform better. For all 200-sample experiments, \ourmodelM turns out to be at least as good as \ourmodel for all batch sizes.

In addition, Tables~\ref{tab:gpu_10} and~\ref{tab:gpu_200} show the performance of \ourmodel and \ourmodelM with respect to different numbers of GPUs used for running the experiments. The reported accuracy for each number is the average accuracy of multiple experiments with different local batch sizes. Again, in most cases \ourmodelM outperforms \ourmodel; in 200-sample experiments, \ourmodelM always performs better than \ourmodel but in 10-sample experiments, \ourmodelM overcomes \ourmodel only when 3 GPUs are used.

\begin{figure*}[ht!]
\hspace*{\fill}%
\includegraphics[scale=.37]{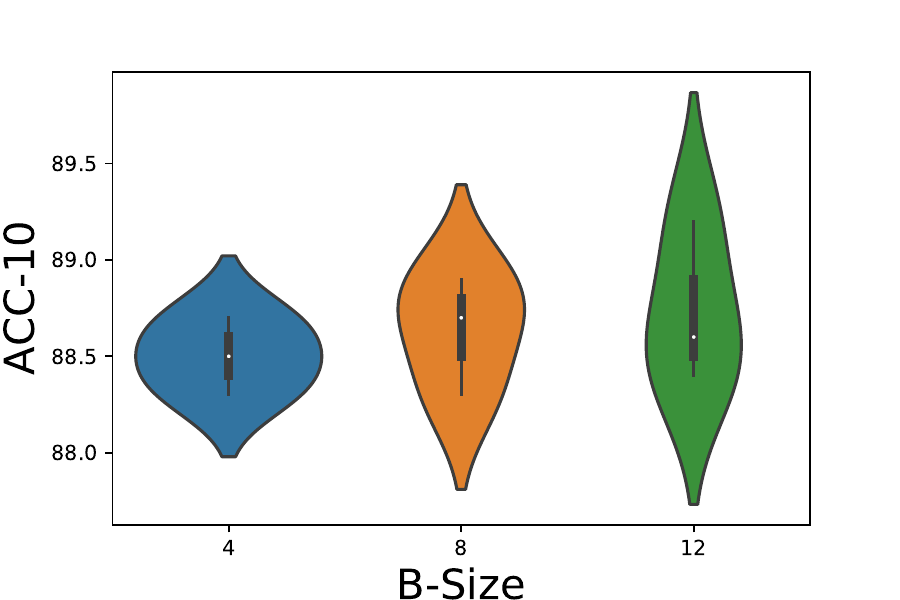}%
\hfill%
\includegraphics[scale=.37]{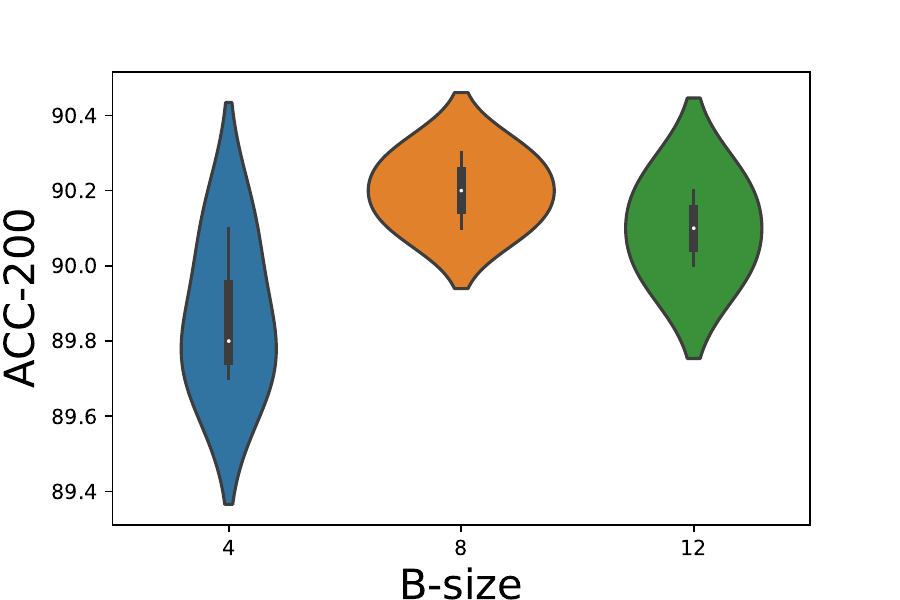}%
\hspace*{\fill}%
\caption{Effect of batch-size per GPU on \ourmodel performance. A local batch size of 4 gives stable performance for the 10-sample case (left), while a batch size of 8 gives better performance for the 200-sample case (right). This suggests that when more training data is available, a larger batch size yields better performance. 
}
\label{fig:violin_charts1}
\end{figure*}

\begin{figure*}[ht!]
\hspace*{\fill}%
\includegraphics[scale=.37]{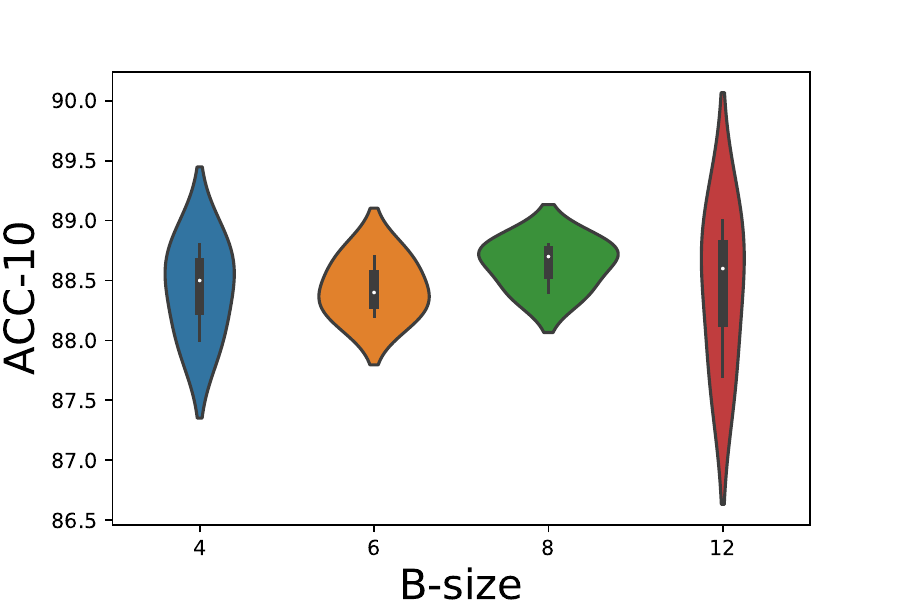}%
\hfill%
\includegraphics[scale=.37]{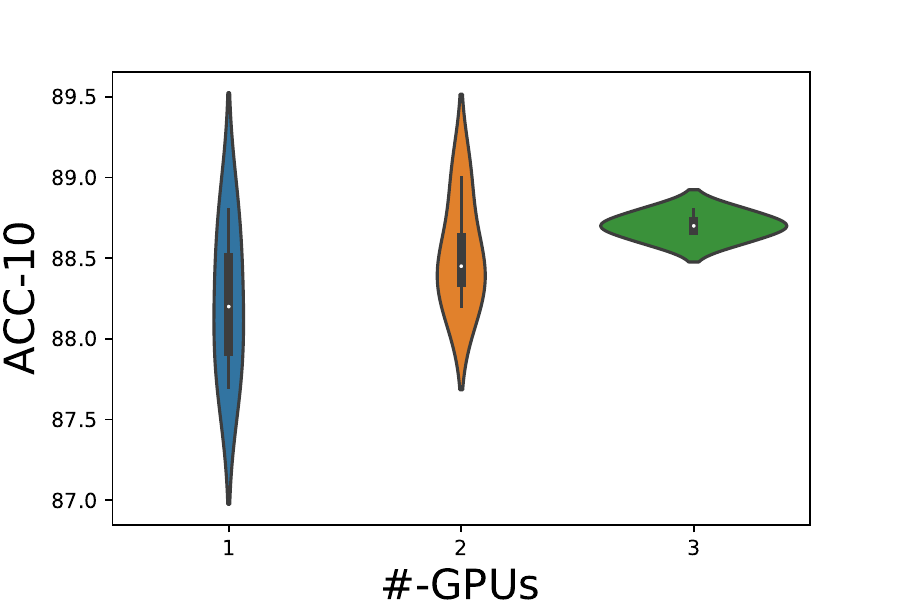}%
\hspace*{\fill}%
\caption{Effect of batch size and number of GPUs on \ourmodelM performance. Similar to \ourmodel, the 10-sample case benefits from a smaller batch size, and \ourmodelM will remain stable (and perform well) with 3 GPUs.}
\label{fig:violin_charts3}
\end{figure*}

We also examined the effect of the local batch size on \ourmodel performance. For this, we ran \ourmodel on different numbers of GPUs using different numbers of local batch sizes. Figure \ref{fig:violin_charts1} illustrates the results of these runs. We observe that when the number of training samples grows, the batch size should be modestly increased (but not too much).
In the case of \ag, Figure \ref{fig:violin_charts1} suggests that for \ourmodel a batch size of 4 gives stable performance for the 10-sample case, while a batch size of 8 gives better performance for the 200-sample case. 
Moreover, Figure \ref{fig:violin_charts3} shows the effect of batch size and number of GPUs on \ourmodelM performance in 10-sample case. Similar to \ourmodel, it benefits from a smaller batch size, and \ourmodelM will remain stable (and perform well) with 3 GPUs.

Overall, based on these experiments, we draw the following conclusions:
\begin{itemize}
    \item When the size of the labeled training data is sufficient (a few hundred samples), \ourmodelM performs better under any choice of the number of GPUs and batch size. 
    \item With a more limited number of supervised data samples, \ourmodelM prefers larger batch sizes than \ourmodel. We hypothesize that 
    this is because \ourmodelM has seen the data before in the pre-training and is thus more stable towards the dataset. Hence, it can be expected to require more coarse-graned information to update itself. 
    \item When the number of training samples grows, a moderate increase in batch size is beneficial for both \ourmodel and \ourmodelM.
\end{itemize}

\begin{table}[t!]
\begin{tabularx}{\linewidth}{@{}XrXXrXXrX@{}}
\toprule
\multicolumn{3}{c}{\multirow{2}{*}{Global B-Size}} & \multicolumn{6}{c}{Accuracy} \\
                              \cmidrule(l){4-9}
&&                              & \multicolumn{3}{c}{\ourmodel}          & \multicolumn{3}{c}{\ourmodelM}          \\
                              \midrule
&4  &                           && \textbf{88.30} &&            & 88.00 &               \\
&6  &                           && \textbf{88.80} &&            & 88.40 &             \\
&8  &                           && \textbf{88.80} &&            & 88.65 &            \\
&12 &                           && \textbf{88.73} &&           & 88.23  &           \\
&16 &                           && \textbf{88.70} &&            & 88.40 &             \\
&18 &                           && \textbf{88.70} &&               & \textbf{88.70}  &            \\
&24 &                           && 88.75          &&   & \textbf{88.85} &            \\
&36 &                           && 88.40          &&   & \textbf{88.60} &             \\
\bottomrule
\end{tabularx}
\caption{The accuracy of \ourmodel and \ourmodelM w.r.t.\ global batch size with 10 labeled samples on \ag\label{tab:batch_10}}
\end{table}%
\begin{table}[t!]
\begin{tabularx}{\linewidth}{@{}XrXXrXXrX@{}}
\toprule
\multicolumn{3}{c}{\multirow{2}{*}{Global B-Size}} & \multicolumn{6}{c}{Accuracy} \\
                              \cmidrule(l){4-9}
&&                              & \multicolumn{3}{c}{\ourmodel}          & \multicolumn{3}{c}{\ourmodelM}          \\
                              \midrule
&4   &                          && 89.70  &&            & \textbf{90.20}  &            \\
&6   &                          && \textbf{90.20} &&                & \textbf{90.20} &             \\
&8   &                          && 90.00 &&               & \textbf{90.15}      &       \\
&12  &                          && 90.20 &&            & \textbf{90.30}       &       \\
&16  &                          && 90.10 &&             & \textbf{90.30}       &       \\
&18  &                          && 90.20 &&                & \textbf{90.40}    &          \\
&24  &                          && 90.20 &&             & \textbf{90.35}      &       \\
&36  &                          && 90.00 &&               & \textbf{90.40}    &          \\
\bottomrule
\end{tabularx}
\caption{The accuracy of \ourmodel and \ourmodelM w.r.t global batch size with 200 labeled samples  on \ag\label{tab:batch_200}}
\end{table}

\begin{table}[t!]
\begin{tabularx}{\linewidth}{cXrXXrX}
\toprule
\multirow{2}{*}{GPUs} & \multicolumn{6}{c}{Accuracy} \\
                       \cmidrule(l){2-7}
                       & \multicolumn{3}{c}{\ourmodel}      & \multicolumn{3}{c}{\ourmodelM}     \\
                       \midrule
1                      && \textbf{88.65} &       && 88.22   &    \\
2                      && \textbf{88.87} &       && 88.52   &    \\
3                      && 88.40          &       && \textbf{88.70}  &       \\
\bottomrule
\end{tabularx}
\caption{The accuracy of \ourmodel and \ourmodelM w.r.t.\ number of GPUs with 10 labeled samples  on \ag\label{tab:gpu_10}}
\end{table}%
\begin{table}[t!]
\begin{tabularx}{\linewidth}{cXrXXrX}
\toprule
\multirow{2}{*}{GPUs} & \multicolumn{6}{c}{Accuracy} \\
                       \cmidrule(l){2-7}
                       & \multicolumn{3}{c}{\ourmodel}      & \multicolumn{3}{c}{\ourmodelM}     \\
                       \midrule
1                      && 90.03  &      && \textbf{90.27}  &      \\
2                      && 89.95  &      && \textbf{90.22}  &      \\
3                      && 90.13  &      && \textbf{90.35}  &       \\
\bottomrule
\end{tabularx}
\caption{The accuracy of \ourmodel and \ourmodelM w.r.t.\ number of GPUs with 200 labeled samples  on \ag\label{tab:gpu_200}}
\end{table}

\subsection{Effect of Topic Word Masking in Zero-Shot Evaluation (\question9)}
Practical applications of topic modeling may aim at topics that are compound or ultra fine grained, that change over time, etc. One approach that can support such application scenarios is to develop models capable of classifying unseen categories without any training instances, so-called zero-shot classification. The rise of pre-trained language models made the idea of providing task descriptions for neural architectures in zero-shot experiments feasible. Therefore, it is a valid question to ask if \ourmodelM performs better in zero-shot classification than \ourmodel.

To study this question, we
\begin{enumerate}
    \item split \yahoo into
    \begin{enumerate*}[label=Yahoo!\ $\textrm{Answers}_{\textrm{(\Alph*)}}$]
    \item\label{YahooA} (society, health, computer, business, relationship) and
    \item\label{YahooB} (science, education, sports, entertainment, politics),
    \end{enumerate*}
    \item pre-train \emph{bert-base-uncased} and \emph{distilbert-base-uncased} language models on \ref{YahooA} using objective masking (see Section \ref{sec:masking}),
    \item train Distil-\ourmodelM on \ref{YahooA} using the language models of the previous step as the text encoders in the teacher and the student models (we use 200 labeled examples and 5\,000 unlabeled examples per class for training),
    \item train Distil-\ourmodel on \ref{YahooA}, and
    \item compare the performance of the two resulting student models on the \ref{YahooB} test set in a zero-shot evaluation setting.
\end{enumerate}
For zero-shot evaluation, we use the Pattern Exploiting Training (PET) approach proposed by \citet{schick-schutze-2021-exploiting} for semi-supervised text classification. Initially, PET trains several language models with labeled data using different input patterns. The ensemble of these language models is then used to predict pseudo labels for unlabeled data. Finally, a standard classifier is trained based on the pseudo-labeled data. In our experiments, we use language models for prediction without training them. So, we combine two language models of the same type with two different input patterns and use them to classify samples in the test set of \ref{YahooB}. Table \ref{tab:patterns-table} shows the cloze style patterns we use for the inputs.

In four different experiments we initialize PET language models with 
\begin{enumerate}
    \item distilbert-base-uncased (DistilBERT)
    \item distilbert-base-uncased pre-trained on \ref{YahooA} with topic word masking (DistilBERTM)
    \item the student model in  Distil-\ourmodel trained on \ref{YahooA} (Student$_{\text{Distil-\ourmodel}}$)
    \item the student model in Distil-\ourmodelM trained on \ref{YahooA} (Student$_{\text{Distil-\ourmodelM}}$)
\end{enumerate}

Table \ref{tab:zero-shot-results} shows the accuracy of PET on the \ref{YahooB} test set in these experiments. 
DistilBERTM and Student$_{\text{Distil-\ourmodelM}}$ outperform DistilBERT and Student$_{\text{Distil-\ourmodel}}$, respectively, which shows that the proposed pre-training of the language model with objective masking can increase the ability of the language model to recognize examples of classes that have not been seen before. Also, the superiority of the student models over the corresponding original models confirms that a knowledge transfer from the teacher to the student happens.

\begin{table}[ht!]
\centering
\begin{tabular}{|ll|}  \hline
\textbf{P1:}&$\langle$Mask$\rangle$ : Text \\
\textbf{P2:}&[Category: $\langle$Mask$\rangle$] Text \\
\hline
\end{tabular}
\caption{\label{tab:patterns-table} Cloze style patterns for zero-shot evaluation. The task of the transformer is to replace $\langle$Mask$\rangle$ with a suitable category label.}
\end{table}

\begin{table}[ht!]
\centering
\begin{tabular}{lc}
\toprule
\textbf{Language Model} & \textbf{Accuracy} \\
\midrule
DistilBERT             & 0.5699 \textpm 0.020    \\
DistilBERTM          & 0.6039 \textpm  0.003    \\
Student$_{\text{Distil-\ourmodel}}$          & 0.5807 \textpm  0.050   \\
Student$_{\text{Distil-\ourmodelM}}$       & \textbf{0.6310 \textpm 0.050}    \\
\bottomrule
\end{tabular}%
\caption{\label{tab:zero-shot-results} Performance of PET with 4 different initializations in the zero-shot evaluation setting.}
\end{table}

\section{Conclusions}
\label{sec:conclusion}

In this paper, we proposed \ourmodelM, an extension of a semi-supervised text classification approach \ourmodel by \citet{Hatefi.etAl:21}. \ourmodelM uses objective masking in an unsupervised pre-training phase to improve \ourmodel. The idea is to use LDA topic modeling for finding lists of words that are likely to carry topic information. 
We adopt relevance scores to select topic words from the LDA topic model. By adjusting the parameter $\lambda$, we can control the specificity of the chosen words. A lower lambda value will prioritize words that are very specific to the topics, while a higher value will include words that are more frequent and may appear in multiple topics. This flexibility allows us to tailor the selection process based on the specific requirements of the dataset. We studied the performance of \ourmodel and \ourmodelM via extensive experiments over three public datasets (\yahoo, \ag, and Medical Abstracts datasets) in English and one private dataset (\bonnier) of news articles in Swedish. While the latter is not publicly available, interested researchers may request access by sending an e-mail to \texttt{\href{mailto:datasets@bonniernews.se}{datasets@bonniernews.se}} in order to reproduce our results or perform their own experiments.

Our experimental findings demonstrate that \ourmodelM outperforms \ourmodel, BERT classifiers (both pre-trained and non-pre-trained), and SoTA baselines in most cases over all datasets. However, the impact of objective masking on classification accuracy is more pronounced when the amount of supervised data for classification is limited. Moreover, \ourmodelM outperforms the variant obtained by using random masking instead of objective masking. However, the effectiveness of objective masking compared to random masking is influenced by dataset characteristics, such as document length, deviation from BERT training data, and the amount of data available for pre-training. For instance, the use of objective masking proves to be particularly effective for the Medical Abstracts dataset, which we believe to be caused by the different characteristics of this dataset compared to the training data used to develop BERT. Additionally, objective masking performs better on \yahoo than on \ag, an effect we attribute to fact that short texts like those of \ag provide less context for BERT during the pre-training phase of \ourmodelM.

Furthermore, our qualitative analysis indicates that pre-training with objective masking can help the language model learn in which contexts -- and thus in which topics -- certain keywords are likely to appear, enabling it to account for contextual factors when classifying documents. This improves the reliability and interpretability of the model and leads to more accurate classification results.


\end{document}